\documentclass[mnsc,nonblindrev]{informs3}
\OneAndAHalfSpacedXI
\usepackage{graphicx}
\usepackage{float}
\usepackage{subfigure}
\usepackage{color}
\usepackage{tabularx}
\usepackage{multirow}
\usepackage{makecell}

\usepackage{CJKutf8}

\usepackage[ruled,linesnumbered]{algorithm2e}
\usepackage{algpseudocode}
\usepackage{hyperref}
\usepackage{natbib}
 \bibpunct[, ]{(}{)}{,}{a}{}{,}%
 \def\bibfont{\small}%
\TheoremsNumberedThrough     
\ECRepeatTheorems

\EquationsNumberedThrough    

\begin{document}
\begin{CJK}{UTF8}{gbsn}



\TITLE{AlphaRank: An Artificial Intelligence Approach for Ranking and Selection Problems}

\ARTICLEAUTHORS{
\AUTHOR{Ruihan Zhou}
\AFF{Wuhan Institute of Artificial Intelligence, Guanghua School of Management, Peking University, Beijing 100871, China; Xiangjiang Laboratory, Changsha 410000, China, \EMAIL{rhzhou@stu.pku.edu.cn}}
\AUTHOR{L. Jeff Hong}
\AFF{School of Management and School of Data Science, Fudan University, Shanghai 200433, China,\EMAIL{hong\_liu@fudan.edu.cn}}
\AUTHOR{Yijie Peng}
\AFF{Wuhan Institute of Artificial Intelligence, Guanghua School of Management, Peking University, Beijing 100871, China; Xiangjiang Laboratory, Changsha 410000, China,\EMAIL{pengyijie@pku.edu.cn}}
} 

\ABSTRACT{We introduce AlphaRank, an artificial intelligence approach to address the fixed-budget ranking and selection (R\&S) problems. We formulate the sequential sampling decision as a Markov decision process and propose a Monte Carlo simulation-based rollout policy that utilizes classic R\&S procedures as base policies for efficiently learning the value function of stochastic dynamic programming. We accelerate online sample-allocation by using deep reinforcement learning to pre-train a neural network model offline based on a given prior. We also propose a parallelizable computing framework for large-scale problems, effectively combining ``divide and conquer" and ``recursion" for enhanced scalability and efficiency.
Numerical experiments demonstrate that the performance of AlphaRank is significantly improved over the base policies, which could be attributed to AlphaRank's superior capability on the trade-off among mean, variance, and induced correlation overlooked by many existing policies.
}

\KEYWORDS{ranking and selection, artificial intelligence, Monte Carlo simulation}
\maketitle

\section{Introduction}\label{section1}
Management decisions often entail the complex task of selecting the best from multiple alternatives in uncertain environments, many of which can be formulated as ranking and selection (R\&S) problems with widespread applications across fields like operations research, engineering, and finance. In R\&S, mean performances are unknown and need to be estimated through statistical sampling. The sampling procedure allocates the observations to each alternative and identifies the best alternative based on the collected sample information. There is a delicate balance between the precision of decision-making and the available simulation budget, typically subject to constraints defined by either fixed-precision or fixed-budget criteria, and the former approach allocates observations to ensure a pre-specified level of probability of correct selection (PCS), whereas the latter aims to maximize PCS (or other metrics) constrained by a given simulation budget.
Besides PCS, the other critical metric for evaluating the final reward is the expected opportunity cost (EOC), which quantifies the economic regret of the selection decision by measuring the performance gap between the chosen and the best alternatives. 

The methods for solving R\&S problems can be categorized into fixed-precision and fixed-budget procedures based on different objectives (\citealt{2012Best}; \citealt{Hunter2017}; \citealt{2021Review}). Fixed-precision procedures are aimed to provide a statistical guarantee that the best (or at least approximately best) alternative will be selected. A well-established indifference-zone (IZ) framework pioneered by \cite{1954A} and \cite{Rinott1978OnTS} introduce a parameter $\delta$ to represent the minimum discernible difference, which can choose either the best alternative or an alternative whose performance is within $\delta$. Recent advances include \cite{2001A}, \cite{2007A}, \cite{2015Efficient}, \cite{2015Fully}, \cite{2021Review}, \cite{2022Knockout} and many others. Fixed-budget procedures are typically designed to intelligently allocate simulation observations to enhance the efficiency of identifying the optimal alternative based on Bayesian decision theory, with the unknown parameters assumed to follow prior distributions and the posterior distributions of the unknown parameters updated according to the Bayesian rule based on the allocated simulation observations, whereas there are few exceptions such as \cite{Hong2022SolvingLF}, which solve the problem from a frequentist view point. \cite{2016Dynamic} and \cite{Peng2018b} introduce a stochastic dynamic programming (SDP) framework to capture the decisions in sequential sampling. However, solving SDP faces challenges in the presence of many sampling steps or many alternatives, which can lead to the curse of dimensionality.
Therefore, the early optimal sample-allocation procedures attempt to solve the problem either formulated as a static optimization or a one-step optimization problem. Some well-known sampling procedures include optimal computing budget allocation (OCBA) (\citealt{2006Efficient}; \citealt{2000Simulation}), expected value of information (EVI) (\citealt{2010Sequential}; \citealt{2001New}), knowledge gradient (KG) (\citealt{2007A}; \citealt{1996Bayesian}), and expected improvement (EI) (\citealt{Donald1998Efficient}; \citealt{0on}). In this paper, we consider how to solve the fixed-budget R\&S problem accurately and fast from the perspective of SDP.

When addressing the fixed-budget R\&S problem through SDP, the computational complexity typically escalates exponentially due to the need for backward induction in calculating the optimal allocation decision for each possible state. Therefore, it is computationally intractable to calculate the optimal sampling policy by traditional SDP techniques. A prevalent approach to address this complexity is through approximate dynamic programming (ADP), which simplifies the problem by approximating the value function instead of calculating it precisely. For instance, ADP often involves optimizing an approximate value function by adopting a one-step look-ahead policy, which, though computationally simpler, might lead to inaccuracies, especially with a large budget. Analogous to the scenario in playing the game of Go, where assessing the probability of winning or losing in the current board position necessitates looking beyond only the next move, because the result is attainable only at the final stage (or at least nearing the final stage) of the game. 

Rollout policy emerges as a promising solution, offering improved estimation of value functions by dynamically refining choices based on simulation outcomes, thus circumventing the computational pitfalls of traditional SDP methods. Rollout policy was first applied by \cite{1996On} in computer backgammon, illustrating its potential in complex decision-making scenarios, and  the terminology ``rollout" is a synonym for repeatedly playing a given backgammon position to estimate the expected score by Monte Carlo simulation. \cite{1999Rollout} formally give the definition of rollout policy in SDP, i.e., starting with a base policy and employing sampling to estimate the value function of the policy, which is then used to make a one-step prediction, leading to the selection of the optimal action. In this paper, we extend the work of \cite{zhou2023}, introduce a rollout policy for learning and decision-making in the fixed-budget R\&S problem, in which the final PCS after selecting one alternative at the current step and then allocating all remaining budgets based on a base policy is regarded as the action-value function of the alternative. Note that the value function in our problem does not have an explicit form that can be calculated, similar to the wining rate of the Go game, the value function can only be estimated by simulation sampling and rollout based on prior information. Since the estimation of action value has noise, the policy improvement is not guaranteed at every step. But even so, the numerical results in this paper still show that rollout policy can significantly improve the final performance of its base policy. Moreover, this paper not only presents theoretical analyses on the likelihood of enhancing base policy performance through rollout policy but also the consistency of the rollout policy. A similar policy for trying to solve the myopic limitation has also been taken by \cite{10015275}, which combines Monte Carlo tree search (MCTS) to design a non-myopic KG procedure. Rollout policy combined with MCTS can look ahead multiple steps, which leads to a more accurate approximate value function but also brings additional computation.

Since each step requires simulation and rollout, the rollout policy is much slower compared with the traditional R\&S method. To mitigate this, one method is to truncate the rollout before the final step, which can accelerate the process but may compromise the accuracy of the action value estimations. Another method is to use an offline-trained neural network (NN) for learning the value functions estimated by the rollout policy and  employing the NN to solve the SDP online, thereby bypassing the need for actual rollout, which can expedite the process while striving to maintain decision quality. Most of the previous researches have studied R\&S problems under the assumption of normality. In this paper, we focus on the R\&S problem under the assumption that both the observations and the unknown mean parameters follow normal distributions. However, our method is not confined to this assumption. We demonstrate the broad applicability of our method in the appendix. Some similar discussions and results can be found in \cite{zhou2023}. Similar to the methodology employed in training image classifiers, we pre-train a NN using a dataset generated based on prior information, a process that enhances the NN's ability to adapt to the dynamics of sampling in R\&S problem, and the NN is then utilized for online sampling in R\&S problem. Rollout policy is used to generate input data, reflecting current states like alternative statistics and remaining budget, and the targets are the estimated action values. The NN learns from this rollout policy, which is iteratively improved by being used as the base policy in the rollout policy in the next training round. The NN's allocation efficiency is evaluated through directly employing the currently trained NN as an allocation policy and assessing whether the resulting PCS meets our predefined stopping rule.

As the number of alternatives increases, the SDP becomes more difficult to solve. Most existing R\&S methods are mainly implemented on problems with the number of alternatives below 100. We prove that the computational complexity of rollout policy is in a square order of the number of alternatives, and the computational complexity of using the NN to guide the allocation is in a linear order of the number of alternatives. Although we propose to pre-train an NN offline for speeding up online sampling, the process requires repeated rollout and training, so it is still difficult to train a large-scale NN. Large-scale R\&S problems often include thousands to millions of alternatives. To enhance computational efficiency, leveraging parallel computing enables the distribution of tasks across multiple processors, significantly speeding up the solution process for large-scale R\&S problems. Recent advancements can be found in \cite{2015Fully}, \cite{2015Efficient}, \cite{Zhong2021SpeedingUP}, \cite{2022Knockout} and \cite{Hong2022SolvingLF}.  The idea  of  ``divide-and-conquer" naturally fits parallel computing platforms that distribute comparison and selection tasks across multiple processors so that solving a large-scale R\&S problem is divided into recursively solving many smaller subproblems. The Knockout-Tournament (KT) and fixed-budget KT (FBKT) procedures introduced by \cite{2022Knockout} and \cite{Hong2022SolvingLF} under the fixed-precision and fixed-budget formulations, respectively, are considered state-of-the-art due to their innovative combination of ``divide-and-conquer” and ``recursion", which effectively streamline the iterative elimination process by pairwise comparison. Building upon this foundation, this paper presents a general parallelizable computing framework called DCR.  The DCR framework outlines a method for decomposing large-scale problems into solvable subproblems of arbitrary scale, which can be addressed recursively.  Any existing fixed-budget R\&S proedure can be applied for solving these smaller subproblems. Numerical results indicate that a well-calibrated balance between model scale and precision, as achieved by using a small, high-precision model, can markedly boost both computational and statistical efficiency in solving large-scale problems within this framework, which makes it more valuable to pre-train small-scale NNs.

Our AI approach, dubbed AlphaRank, draws inspiration from AlphaGo Zero (\citealt{2}), particularly in its use of deep learning techniques and strategic decision-making frameworks, though adapted for the unique challenges of R\&S problems. AlphaRank's objective is to develop a NN model proficient in making allocation decisions, achieved through targeted pretraining that incorporates specific prior distribution data. Different from AlphaGo Zero, AlphaRank does not use the policy network to estimate the optimal policy but only uses the value network to estimate the action value. AlphaRank introduces a revolutionary AI-powered computing paradigm, bringing advancements in efficiency and decision accuracy to the field of R\&S problem-solving. While the offline pre-training is resource-intensive, its one-time training process yields models that can be repetitively utilized online for R\&S problems, offering a significant long-term computational advantage. AlphaRank demonstrates exceptional versatility, allowing for the training of NN models adaptable to a wide range of priors and differing numbers of alternatives, showcasing superior performance compared to existing R\&S procedures. Furthermore, the integration of these efficiently trained small-scale NN models with the DCR framework opens avenues for effectively tackling large-scale R\&S problems, a significant stride in the field.

Although Go is an ancient board game played for centuries, AlphaGo (\citealt{5}) and AlphaGo Zero (\citealt{2}) defeated the best human players, and overturned many authoritative rules in Go playbooks, which are the crystallization of human wisdom over centuries. Analogously, R\&S problems have been studied actively in past several decades, but we show that using AlphaRank can not only significantly elevate the standard of R\&S procedures but also provide new insights into how to design a well-performed procedure by analyzing its sampling behavior in low-confidence scenarios that have been relatively underexplored in the R\&S literature. A pivotal finding by \cite{8118094} reveals that PCS does not always increase monotonically with the simulation budget, a counterintuitive phenomenon that underscores the complexity of R\&S problem-solving, particularly in scenarios with large variances and small mean differences, which is due to the decrease of the correlations induced by the variance of the best alternative when it is allocated additional observations. Specifically, PCS may decrease when additional observations are allocated to the best alternative in a low-confidence scenario. Almost all existing R\&S procedures favor allocating more simulation observations to the best alternative, overlooking induced correlations that are less significant in the frequently investigated high-confidence scenarios but important in under-researched low-confidence scenarios (\citealt{7408526}). Numerical results in Section \ref{section8} reveal that our proposed AlphaRank can avoid the non-monotonicity of PCS in low-confidence scenarios, and the sampling ratio of AlphaRank is rather different from that of the asymptotically optimal sampling ratio in low-confidence scenarios, but they tend to be consistent in high-confidence scenarios. We conjecture that AlphaRank's superior ability to achieve a higher PCS might stem from its capability on the trade-off among mean, variance, and induced correlation which has been neglected by most existing procedures.

 The rest of this paper is organized as follows. In Section \ref{section2}, we introduce the problem formulation and the MDP modeling for fixed-budget R\&S problem under the setting where the observations are normally distributed with known variance. In Section \ref{section3}, we introduce the rollout policy and conduct the theoretical analysis on its probability lower bound of policy improvement at each step, consistency and computional complexity. Then, we design the pre-training process where the rollout policy serves as a tool for policy improvement in Section \ref{section4}. In Section \ref{section5}, we further propose the DCR framework to make the procedure more suitable for solving large-scale R\&S problems in parallel computing environments. We provide the description of the AlphaRank algorithm and the detailed setup for its use to the practical R\&S problems in Section \ref{section6}. In Sections \ref{section7} and \ref{section8}, we conduct comprehensive numerical experiments to test the performance of our procedures and compare them with existing procedures, which demonstrates the potential superiority of our procedures on the trade-offs among mean, variance, and induced correlation. Section \ref{section9} provides conclusions. Proofs and some additional discussions are included in the appendix.

\section{Fixed-budget R\&S as MDP}\label{section2}
In this section, we first state the mathematical formulation for the fixed-budget R\&S problem considered in this paper, and then model it as an MDP.
\subsection{Formulation}\label{section21}
We capture the statistics of a set of alternatives indexed by $i=1,2,\dots,N$, and observe a sequence of data $X_{i,1}, X_{i,2},\dots, X_{i,t}$, where $X_{i,t}$ is the observation at time step $t$ for alternative $i$. Note that the $t$-th index of the observations of alternative 
$i$ may be different from the $t$-th index of steps, because the $i$-th alternative is not necessarily allocated at each step. We use Assumption \ref{ass1} to describe the structure of the R\&S problem in the special case of normal distribution with conjugate prior, which is one of the most common assumptions under the Bayesian framework (\citealt{Gelman2010BayesianDA}).

\begin{assumption}\label{ass1}
Suppose $X_{i,t}\sim N(\mu_i^{true}, (\sigma_i^{true})^2)$, where parameter $\mu_i^{true}$ is unknown and $(\sigma_i^{true})^2$ is known, and $\mu_i^{true}$ follows a normal conjugate prior, i.e., $\mu_i^{true}\sim N(\mu_i^{(0)},(\sigma_i^{(0)})^2)$.
\end{assumption}
To simplify the problem, we assume that there is only one best alternative, whereas the following paradigm is equally valid for problems with multiple best alternatives. Our objective is to find the best alternative defined by $\arg\max_{i=1,\dots, N}\mu_i^{true}$. Let $s_t$ be the current state, which contains complete environment information, including the statistics $\varepsilon_t$ of the observations. Let the selection after allocating $t$ simulation observations be $\widehat{S}_t$. Then, the PCS given the current state $s_t$ is 
\begin{gather}\label{eqpcs}
    \text{PCS}(s_t)\triangleq\text{Pr}\left( \left.\widehat{S}_t=\arg\max_{i=1,\dots,N}\mu_i^{true}\right|s_t\right).
\end{gather}
\subsection{MDP Modeling}\label{section22}
For the fixed-budget R\&S problem, the statistics $\varepsilon_t$ of the alternatives at step $t$ under Assumption \ref{ass1} could be 
\begin{equation}\label{eq1}
    \varepsilon_t= \Big\{\Bar{X}_t, \Bar{\sigma}_t^2, \mu^{(t)}, (\sigma^{(t)})^2\Big\},
\end{equation} 
where
$\Bar{X}_t=\{\Bar{X}_{1,t},\dots,\Bar{X}_{N,t}\}$, $\Bar{\sigma}_t^2=\{\Bar{\sigma}_{1,t}^2,\dots,\Bar{\sigma}_{N,t}^2\}$, $ \mu^{(t)}=\{\mu_{1}^{(t)},\dots,\mu_{N}^{(t)}\}$, and $ (\sigma^{(t)})^2=\{(\sigma_1^{(t)})^2,\dots,(\sigma_N^{(t)})^2\}$. The statistics $\Bar{X}_{i,t}$, $\Bar{\sigma}_{i,t}^2$, $\mu_{i}^{(t)}$, and $(\sigma_i^{(t)})^2$ are the sample mean, sample variance, parameter prior mean, and parameter prior variance of alternative $i$, respectively. 

We set the total simulation budget at a fixed value of $T>0$. At step $t$, $t=1,2,\dots,T-1$, there are $t$ simulation observations that have been allocated, so the remaining simulation budget is $T-t$. Therefore, the state at step $t$ is
$$s_t=\{\varepsilon_t, T-t\}=\Big\{\Bar{X}_t, \Bar{\sigma}_t^2, \mu^{(t)}, (\sigma^{(t)})^2, T-t\Big\}.$$
In this simulation optimization setting, the primary objective is to maximize a key objective metric, such as PCS, subject to simulation budget constraint. This can be achieved by computing an optimal policy, which specifies the action to take in each state for maximizing the expected reward given the remaining budget. The state that can be achieved after allocating the $(t+1)$-th simulation observation to alternative $i$ is 
\begin{gather}
\label{eq2}
s_{t+1}^{(i)}=\left\{\varepsilon_{t+1}^{(i)}, T-t-1\right\},
\end{gather}
where $\varepsilon_{t+1}^{(i)}$ represents the updated statistics after the $(t+1)$-th simulation observation has been allocated to alternative $i$.
The fixed-budget R\&S problem as MDP can be characterized by the following Bellman equation. For $0\leq t<T$, we have
$$a_{t+1}^*=\arg\max_{i=1,\dots,N}Q(s_{t},i),$$
$$V_t(s_t)=Q\left(s_{t},a_{t+1}^*\right)=\mathbb{E}\left[Q\left(\left.s_{t},a_{t+2}^*\right)\right|s_t, a_{t+1}^*\right]=\mathbb{E}\left[\left.V_{t+1}(s_{t+1})\right|s_t, a_{t+1}^*\right],$$
where $a^*_{t+1}$ is the optimal allocation policy at $t+1$-th step, $Q(s_{t}, i)$ is the state-action-value function, which represents the reward value of choosing alternative $i$, and $s_{t+1}$ is determined by $s_t$ and newly allocated observation $X_{a^*_{t+1},t+1}$. In principle, the action values $Q(s_{t}, i)$ can be calculated using classic SDP techniques, such as value iteration or policy iteration. Many value-function approximations have been proposed in existing R\&S studies, such as KG, OCBA, and AOAP. The posterior PCS is calculated by (\ref{eqpcs}).

To identify the best alternative from the set of $N$ alternatives based on the observations, the alternative that maximizes the posterior PCS, based on the information of all allocated simulation observations, is selected, given the current state. 
For PCS, the optimal selection policy at the state $s_T$ of step $T$ is (\citealt{2016Dynamic})
\begin{gather}\label{eq2122}
    S_T^*= \arg \max_{i=1,\dots,N}  \text{Pr}\left(\left.\mu_i^{true}\geq\mu_j^{true}\right|s_T\right),
\end{gather} 
and the value function after allocating $T$ simulation observations is 
$$V_T(s_T)=Q(s_T, S_T^*)=\mathbb{E}\left[\left.V\left(s_T, S_T^*\right)\right|s_T\right]=\mathbb{E}\left[\left.\textbf{1}\left\{ S_T^*=\arg\max_{i=1,\dots,N}\mu_i^{true}\right\}\right|s_T\right].$$
In particular, the approximate optimal selection policy in the normal-conjugacy case is the largest posterior mean, i.e., 
\begin{equation}\label{eqselect1}
\widehat{S_{T}}=\arg\max_{i=1,\dots,N}\mathbb{E}\left[\left.\mu_{i}^{true}\right|s_T\right]=\arg\max_{i=1,\dots,N}\mu_{i}^{(T)},
\end{equation} which is often used to simplify the calculation of (\ref{eq2122}) (\citealt{1954A}).

Following each sampling, the current state is updated based on the observations and the remaining simulation budget.
 In the current discussion, the state transition mechanism is to update $\Bar{X}_{t+1}$, $\Bar{\sigma}_{t+1}^2$, $\mu^{(t+1)}$, and $(\sigma^{(t+1)})^2$ according to the latest allocated observation. Let $T_{i,t}$ represent the number of simulation observations allocated to alternative $i$ with a total of $t$ allocated simulation observations, $\sum_{i=1,\dots,N}T_{i,t}=t$. If the $(t+1)$-th simulation observation is allocated to alternative $i$, then $T_{i,t+1}=T_{i,t}+1$ and the observation is $X_{i,t+1}$. Specifically, the sample mean $\Bar{X}_{i,t+1}$ and sample variance $\Bar{\sigma}_{i,t+1}^2$ of alternative $i$ are updated as follows:
$$\Bar{X}_{i,t+1}=\frac{T_{i,t}\cdot\Bar{X}_{i,t}+X_{i,t+1}}{T_{i,t+1}},\ \Bar{\sigma}_{i,t+1}^2=\frac{T_{i,t}}{T_{i,t+1}}\cdot\left(\Bar{\sigma}_{i,t}^2+\frac{(\Bar{X}_{i,t}-X_{i,t+1})^2}{T_{i,t+1}}\right).$$
Under Assumption \ref{ass1}, $ N(\mu_i^{true}, (\sigma_i^{true})^2)$ with an unknown mean and known variance is another normal distribution $N(\mu_i^{(0)},(\sigma_i^{(0)})^2)$. The posterior distribution of $\mu_i^{true}$ is $N(\mu_i^{(t)},(\sigma_i^{(t)})^2)$, where
$$
    \mu_i^{(t)}=(\sigma_i^{(t)})^2\left(\frac{\mu_i^{(0)}}{(\sigma_i^{(0)})^2}+\frac{T_{i,t}\cdot\Bar{X}_{i,t}}{(\sigma_i^{true})^2}\right),\ (\sigma_i^{(t)})^2=\left(\frac{1}{(\sigma_i^{(0)})^2}+\frac{T_{i,t}}{(\sigma_i^{true})^2}\right)^{-1}.
$$
If $(\sigma_i^{(0)})^2\to \infty$, then $\mu_i^{(t)}=\Bar{X}_{i,t}$. When the parameter distribution does not satisfy conjugacy, we propose to update the posterior distribution by using the particle filtering method. We describe the method and give the corresponding numerical experimental results in Appendices \ref{app:general} and \ref{app:gamma}. After getting  $\mu_i^{(t)}$ and $(\sigma_i^{(t)})^2$, we predict the distribution of the observations for alternative $i$ is $N(\mu_i^{(t)},(\sigma_i^{true})^2)+(\sigma_i^{(t)})^2)$. Parameters of other alternatives except for alternative $i$ being allocated will not be updated at the $(t+1)$-th step.

\section{Rollout Policy}\label{section3}
In this section, we first introduce the rollout policy for solving fixed-budget R\&S problems. We then show the probability lower bound of policy improvement at each sampling step and prove the rollout policy is consistent. Finally, we analyze its computational complexity. Note that all discussions herein utilize PCS as the metric. Correspondingly, the reward functions and the design of policies are also derived based on PCS. However, it is feasible to design policies and algorithms based on other metrics such as EOC, and the theoretical properties and insights of these policies remain consistent regardless of the specific metric employed.
\subsection{Procedure}\label{section31}
We first introduce the key notation used in Section \ref{section3}.
\begin{itemize}
    \item $a_{t+1}^{(i)}$ is the possible action that allocates the $(t + 1)$-th simulation observation to alternative $i$, whose action value is estimated by rollout policy. The set of possible actions for allocating the $(t + 1)$-th simulation observation is $\{a_{t+1}^{(1)},\dots, a_{t+1}^{(N)}\}$.
    \item $s_{t+1}^{(i)}$ is the updated state after selecting the $i$-th alternative through action $a_{t+1}^{(i)}$ at current state $s_{t}$. Refer to (\ref{eq2}) for the state transition. 
    \item $Q(s_t,a_{t+1}^{(i)})$ is the theoretical action value of taking action $a_{t+1}^{(i)}$ in $s_{t}$. In our rollout policy, it represents the PCS value $\text{PCS}^{\pi}(s_{t+1}^{(i)})$ that can be obtained when the remaining simulation observations are allocated through the base policy $\pi$ after action $a_{t+1}^{(i)}$ is selected at $s_{t}$. Since it is difficult to have an explicit form to calculate $Q(s_t,a_{t+1}^{(i)})$ precisely, we use an approximation method to estimate this value.
    \item $Q_t^{(i)}(s_{t+1}^{(i)})$ is an approximation of $Q(s_t,a_{t+1}^{(i)})$, which is estimated from the average of the instant rewards $r_{t,1}^{(i)}, r_{t,2}^{(i)}, \dots, r_{t,k}^{(i)}$ over $K$ rollouts.
\end{itemize}

The idea of the rollout policy with a base policy may be stated as follows: given a base policy $\pi$, perform $K$ rollouts with $\pi$ at the $i$-th decision node at step $t$ (i.e., selecting the $i$-th alternative to allocate the $(t+1)$-th simulation observation by the action $a_{t+1}^{(i)}$), and estimate the action value of $a_{t+1}^{(i)}$, i.e., $Q_t^{(i)}(s_{t+1}^{(i)})$. Upon estimating the action value, the $(t+1)$-th simulation observation is then strategically allocated to the alternative that exhibits the highest calculated action value at that point, i.e.,  the sample-allocation action of the rollout policy is
 \begin{gather}\label{eq311}
a_{t+1}^{roll}=\arg\max_{i=1,\dots,N}Q_t^{(i)}\left(s_{t+1}^{(i)}\right).
 \end{gather}
 The selection after all $T$ simulation observations have been allocated is 
 \begin{align}\label{rolloutwithp1}
S_T^{roll}\left(s_{T}^{(a_{T}^{roll})}\right)=\arg \max_{i=1,\dots,N} \text{Pr}\left(\left.\mu_i^{true}\geq\mu_j^{true}\right|s_T\right),
 \end{align}
For simplicity, we can also approximately select the alternative with the largest posterior
mean, i.e., $S_T^{roll}(s_{T}^{(a_{T}^{roll})})=\arg\max_{i=1,\dots,N}\mathbb{E}\left[\left.\mu_{i}^{true}\right|s_T\right].$

Next, we introduce in detail how to use rollout to calculate $Q_t^{(i)}(s_{t+1}^{(i)})$ by Monte Carlo simulation. Note that the observations used in rollout are not sampled from the simulation models but sampled from the current updated prior. Therefore, they are computationally much easier to obtain and they may be obtained offline where the simulation models may not even exist. This is a critical point that we will take advantage of in pre-training introduced in Section \ref{section4}.

The rollouts entail the generation of $K$ trajectories by the base policy $\pi$, which means that starting from $s_{t+1}^{(i)}$, the remaining $T-t-1$ budget is allocated according to $\pi$. Then we can get  the rewards $r_{t,k}^{(i)}$.  In the $k$-th rollout, $r_{t,k}^{(i)}=1$ when the selection is correct, and $r_{t,k}^{(i)}=0$, otherwise, with probabilities $\text{PCS}^{\pi}(s_{t+1}^{(i)})$ and $1-\text{PCS}^{\pi}(s_{t+1}^{(i)})$, respectively, i.e., $r_{t,k}^{(i)}$ follow the i.i.d. Bernoulli distribution with parameter $\text{PCS}^{\pi}(s_{t+1}^{(i)})$. The aggregate number of correct selections at state $s_{t+1}^{(i)}$ after $K$ rollouts, denoted as $R_{t, K}^{(i)}=\sum_{k=1}^K r_{t,k}^{(i)}$, follows a binomial distribution, i.e., $R_{t, K}^{(i)}\sim B(K,\text{PCS}^{\pi}(s_{t+1}^{(i)}))$. Therefore, $Q_t^{(i)}(s_{t+1}^{(i)})$ can be calculated by $$Q_t^{(i)}\left(s_{t+1}^{(i)}\right)=\frac{1}{K}\sum_{k=1}^K r_{t,k}^{(i)}=\frac{1}{K}R_{t, K}^{(i)}.$$
As $K\rightarrow \infty$, $Q_t^{(i)}(s_{t+1}^{(i)})\rightarrow Q(s_t,a_{t+1}^{(i)})=\text{PCS}^{\pi}(s_{t+1}^{(i)})$. However, due to limited computational resources, the number of rollouts $K$ at each step is bound to be limited, which means that it is likely that the estimate of $Q(s_t,a_{t+1}^{(i)})$ contains noise. 

When the simulation budget $T$ is large, it is computationally expensive to rollout till the end. To address the computational limitation, we can use a rolling horizon rollout policy, which involves a limited number of steps from the current state to the future. Unlike the global rollout for fixed-budget R\&S problems, which simulates $T-t$ steps forward at step $t$, the rolling horizon rollout policy simulates fixed $H$ steps forward at every step, so the estimated value $Q_t^{(i)}(s_{t+1}^{(i)})$ there is used to represent the PCS that can be achieved after $H$ simulation observations are allocated at step $t$. Therefore, it also makes the estimate of $Q(s_t,a_{t+1}^{(i)})$ contain noise. We will analyze the properties of the rollout policy in the case that $Q(s_t,a_{t+1}^{(i)})$ is not accurate enough in the following Sections \ref{section32} and \ref{section33}.

\begin{figure}[htbp]
    \centering
    \includegraphics[width=1\textwidth]{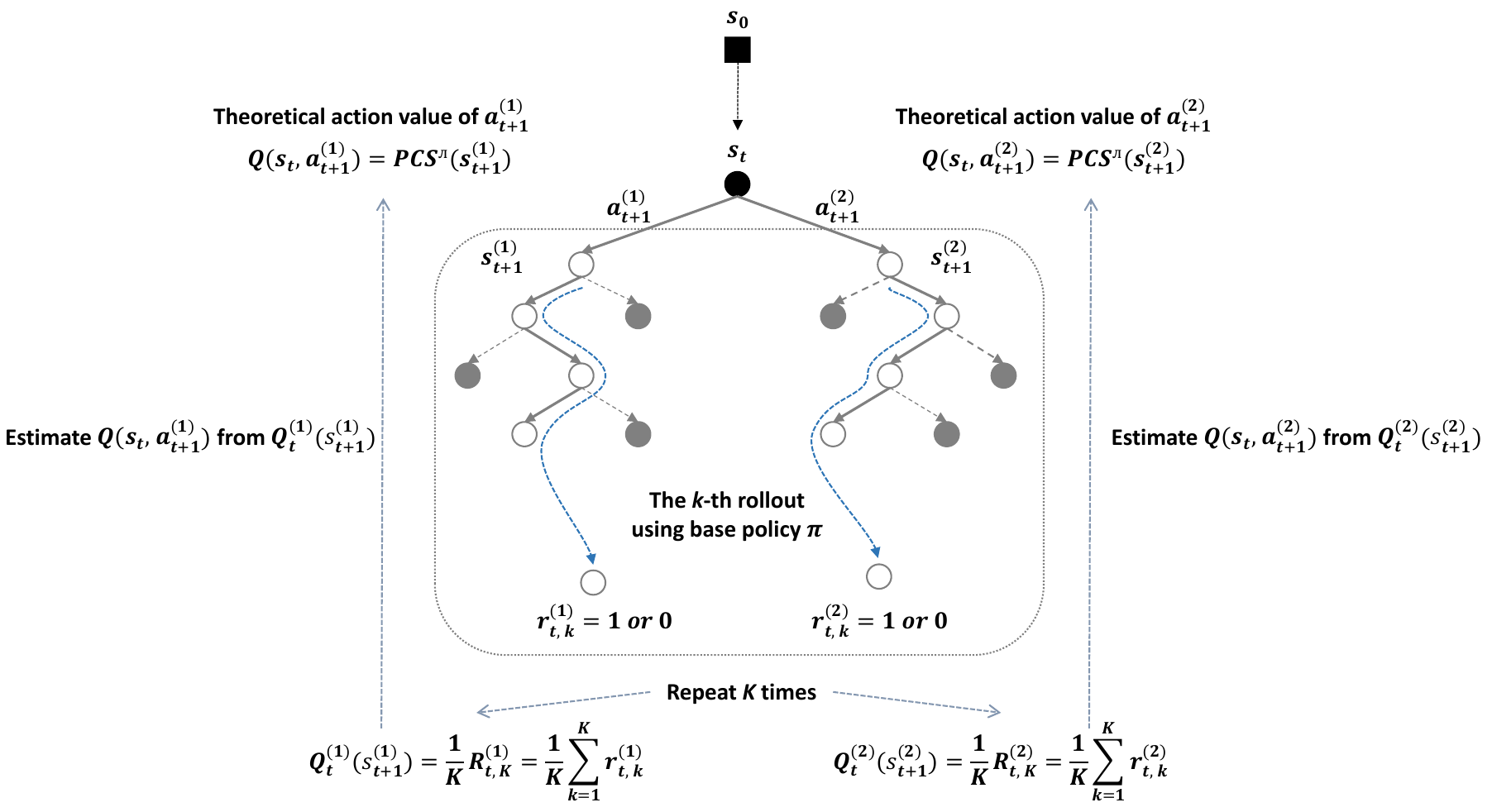}
    \caption{The decisions in the rollout process.}
    \label{rollout}
\end{figure}

 The diagram of the rollout process is shown in Figure \ref{rollout}, which illustrates the decision-making process and state transitions in the rollout, using an example with 2 alternatives to demonstrate how each allocation decision affects the state of the alternatives. In Figure \ref{rollout}, distinct circle types denote different states: black circles indicate alternatives that have not been allocated any simulations, hollow circles signify alternatives currently being allocated simulations, and gray circles represent the simulation observations (i.e., samples from the updated prior distributions) within the rollout process.

\subsection{The Lower Bound of Policy Improvement}\label{section32}
In this subsection, we introduce the following assumption to provide a foundation for our justifications. Notably, the existing R\&S procedures used as base policies, including KG, OCBA, AOAP, among others, are all consistent, meaning they reliably produce increasingly accurate results over time.
\begin{assumption}\label{ass5}
    The base policy $\pi$ of the rollout policy is consistent.
\end{assumption}
According to \cite{1999Rollout}, under certain conditions, it is guaranteed that the rollout policy will perform at least as well as its base policy. However, \cite{1999Rollout} come to their conclusion when the value function has an explicit form. A critical issue arises because the action value of the rollout policy in our scenario is derived from Monte-Carlo estimation, which, due to its reliance on a finite number of rollouts (i.e., $K$) and the fact that sometimes the rollout does not extend to the end, introduces noise and potential inaccuracies in the estimations of $Q_t^{(1)}(s_{t+1}^{(1)}), 
\dots, Q_t^{(N)}(s_{t+1}^{(N)})$. This implies the ordering of estimated values may not always align with their theoretical counterparts $Q(s_t, a_{t+1}^{(i)})$, which can impact the reliability of policy decisions. Only when the estimated values under the current state have correct comparisons with theoretical values, i.e., if $Q(s_t, a_{t+1}^{(i)})>Q(s_t, a_{t+1}^{(j)})$, then $Q_t^{(i)}(s_{t+1}^{(i)})>Q_t^{(j)}(s_{t+1}^{(j)})$, can we guarantee policy improvement at current step. In the next proposition, we show that,  when the estimated action values have noise, we can still guarantee policy improvement with a certain probability. The proof can be found in Appendix \ref{proof2}.

\begin{proposition}\label{pro1}
The rollout action value of action $a_{t+1}^{(i)}$ in state $s_t$ is the theoretical PCS of the base policy $\pi$ in state $s_{t+1}^{(i)}$, i.e., $Q(s_t,a_{t+1}^{(i)})=\text{PCS}^{\pi}(s_{t+1}^{(i)})$. Furthermore, define ${\rm Pr}^{improve}\left(s_t\right)$ as the probability of policy improvement based on the rollout policy in $s_t$. Then, when the number of rollouts used for value function estimation is $K$, we have
\begin{align}\label{pro1eq}
{\rm Pr}^{improve}\left(s_t\right)\geq1-\sum_{i=1}^N\left(1- {\rm Pr}\left(\left.Q_t^{(i)}(s_{t+1}^{(i)})-Q_t^{(a^*)}(s_{t+1}^{(a^*)})\leq 0\right|s_t\right)\right),
\end{align}
where $a^*=\arg\max_{i=1,\dots,N}Q(s_t,a_{t+1}^{(i)})$. 
\end{proposition}
\begin{remark}
Here, ``policy improvement'' is defined as the scenario in which a rollout policy, based on an existing R\&S algorithm, outperforms the direct application of that algorithm (e.g., achieving a higher PCS). Proposition \ref{pro1} shows that with limited number of rollouts, the probability of policy improvement guarantee is strongly related to the PCS of the base policy, because the base policy with better performance can better estimate the action value $Q_t^{(i)}(s_{t+1}^{(i)})$, so that the estimated value $Q_t^{(a^*)}(s_{t+1}^{(a^*)})$ of the true optimal alternative is more likely to be the largest, that is, ${\rm Pr}(Q_t^{(i)}(s_{t+1}^{(i)})-Q_t^{(a^*)}(s_{t+1}^{(a^*)})\leq 0|s_t)$ is greater. Consequently, the effectiveness of the base policy directly influences the rollout policy's performance, and a higher PCS in the base policy typically leads to a higher PCS in the rollout policy. 
\end{remark}

When $K$ is large enough, $R_{t,K}^{(i)}$ approximately follows a normal distribution. Since $R_{t, K}^{(1)}-R_{t, K}^{(a^*)},\dots, R_{t, K}^{(N)}-R_{t, K}^{(a^*)}$ are positively correlated, $i,j=1,2,\dots, N$, then they approximately follow a multivariate normal distribution with positive correlations (\citealt{2016Dynamic}), and Slepian's inequality can be applied to give an approximate lower bound for the probability of policy improvement, which is typically tighter than the probability lower bound (\ref{pro1eq}) obtained by the Bonferroni's inequality. 
\subsection{Consistency}\label{section33}
As highlighted earlier, uncertainty in each step of the rollout policy's allocation can lead to discrepancies between estimated and theoretical action values, potentially affecting accuracy. Given this, it becomes critical to assess whether the rollout policy maintains its consistency in the face of such uncertainty, specifically, we must determine if the policy reliably identifies the true optimal alternative when the budget $T$ becomes very large, approaching infinity. To establish this consistency, we have identified two key assumptions, under which the rollout policy's consistency has been demonstrated.
\begin{assumption}\label{ass6}
    There exists $\epsilon>0$ such that $\epsilon<\text{Pr}(\mu_i^{true}>x|s_T)<1-\epsilon$ for $i=1,2,\dots, N$, any $x$, and any finite $t_i$, where $t_i$ is the number of times alternative $i$ is sampled.
\end{assumption}
\begin{assumption}\label{ass7}
When budget $t\rightarrow \infty$, the probability of policy improvement at step $t$ is positive, i.e., ${\rm Pr}^{improve}\left(s_t\right)>0$.
\end{assumption}
Assumption \ref{ass6} posits that if the number of samples allocated to a non-optimal alternative is finite (i.e., does not approach infinity), then the sample variance for that alternative cannot converge to zero. This leads to a non-negligible probability of selecting such an alternative, thereby hindering the PCS from attaining the ideal value of 1, i.e., if $t_i<\infty$, then $\text{PCS}^{roll}(s_t^{(a_t^{roll})})<1$. Assumption \ref{ass7} indicates there is a positive probability that $\text{PCS}^{roll}(s_t^{(a_t^{roll})})\geq\text{PCS}^{\pi}(s_t^{(a_t^{\pi})})$, where $a_{t}^{roll}$ and $a_{t}^{\pi}$ represent the sample-allocations of the rollout policy and $\pi$, and $\text{PCS}^{roll}$ and $\text{PCS}^{\pi}$ represent the PCS obtained based on $a_{t}^{roll}$ and $a_{t}^{\pi}$, respectively. 

Building on these assumptions, we present the following proposition: the rollout policy’s final selection is expected to converge to the optimal solution as the simulation budget approaches infinity. The proof of the proposition can be found in Appendix \ref{proof3}.
\begin{proposition}\label{pro2}
  Define the selection given by the rollout policy after allocating $T$ simulation observations as $S_T^{roll}(s_{T})$. If Assumptions \ref{ass6} and \ref{ass7} are satisfied, the rollout policy is consistent, i.e.,
 $$\lim_{T\rightarrow \infty}S_T^{roll}(s_{T})=\arg\max_{i=1,\dots,N}\mu_i^{true},\ a.s.$$
\end{proposition}
\subsection{Complexity Analysis}\label{section34}
In addressing the R\&S problem involving $N$ alternatives and a total simulation budget $N$, Proposition \ref{pro_complex1} aims to elucidate the computational complexity associated with the rollout policy.
 \begin{proposition}\label{pro_complex1}
     Suppose the maximum number of the steps in the rollout is $H$. Then the computational complexity of the rollout policy is $O(N^2TH)$.
 \end{proposition}
 \begin{remark}
 With a smaller simulation budget $T$, it is feasible to simulate the remaining $T-t$ steps in the rollout at each time step, resulting in a polynomial computational complexity, specifically $O(N^2T^2)$. However, when $T$ is large, simulating until the end at each step becomes computationally intractable. To mitigate this, setting a fixed number of forward steps $H$ is necessary, which, in turn, brings the complexity down to a linear relationship with $T$, represented as $O(N^2T)$. Furthermore, the computational complexity of the rollout policy in relation to the number of alternatives $N$ is also polynomial.  Consequently, the direct application of the rollout policy in addressing large-scale problems could encounter practical difficulties and efficiency issues.
 \end{remark}

\section{Pre-training Process}\label{section4}
To address the computational efficiency challenges associated with online sampling in the rollout policy,  we propose developing an NN model that emulates the behavior of the rollout policy, achieved through comprehensive offline pre-training guided by a specified prior distribution, and then the trained NN model can be used for guiding online sampling. In this section, we first introduce the pre-training procedure, then we analyze the complexity of NN training and using NN directly as the allocation policy, and offer techniques to speed up the training process.
\subsection{Procedure}\label{section41}
The primary objective of the pre-training procedure is to develop a NN model that demonstrates exceptional performance in allocation tasks.  Initially, it is essential to clearly define the input, output, and training objectives of the NN. Subsequently, appropriate labeled data must be selected for use in the training process. Finally, it is necessary to establish reasonable evaluation criteria to quantify the performance of the NN model after training. The following will elaborate on the process of the specific training methods for NNs, the generation of training data, the evaluation metrics for assessing the performance of the trained NN, and the criteria for determining the stopping point of training.

During training, the NN model is designed to output the estimated action value for each alternative, which is determined based on the input data reflecting the current state. To be specific, the input of NN $input_t$ at state $t$ includes the statistics of the alternatives and the remaining budget, i.e.,
$$input_t=\{\varepsilon_t, T-t\}.$$ 
For example, in the normal-conjugate case we discuss in this paper, the input could be $input_t=\{\Bar{X}_t, \Bar{\sigma}_t^2, \mu^{(t)}, (\sigma^{(t)})^2, T-t\}$, which has 4 sample statistics including sample mean, sample variance, parameter prior mean and variance, and each statistic is an $N$-dimensional vector which is described in Section \ref{section21}. 
When each rollout does not complete the entire set of remaining $T-t$ steps, it is essential to adjust the remaining budget in the input to reflect the actual number of forward steps taken in the rollout. The output of NN $output_t$ is an action value vector $$output_t= V_t=\left(V_t^{(1)},\dots, V_t^{(N)}\right),$$ which evaluates the expected PCS for performing each action at the current state $t$, where $V_t^{(i)}$ represents the action value of selecting alternative $i$ at step $t$. 

Since the NN model is trained to approximate the value function of the rollout policy, the objective of training is to optimize the loss between the NN predicted value $V_t$ and the rollout action value $Q_t$, where $Q_t=(Q_t^{(1)},\dots, Q_t^{(N)})$. We utilize the cross-entropy with a regularization term as the loss function for training NN, which is consistent with the form used by AlphaGo Zero to train the value NN (\citealt{2}):
$$ Loss=-\frac{1}{N}\sum_{i=1}^N\left[V_t^{(i)}\log Q_t^{(i)}+(1-V_t^{(i)})\log (1-Q_t^{(i)})\right]+c||\cdot||^2,$$
where the regularization term $c||\cdot||^2$ is used to prevent the model from overfitting.  Parameters for the setup of the NN in this paper are shown in Appendix \ref{APPD:PARA}. 

Since $N$, the scale of practical problems, may be unknown, and considering pre-training small-scale NNs is relatively easy, we propose training a series of small-scale NNs, ranging in scale from 2 to 10, under a designated prior, to address this issue. For instances where the practical problem scale exceeds the maximum scale of our trained NNs, we offer a solution method in Section \ref{section5}.

The process for generating training data in each round of the training involves several steps. First, a large number of observations is generated based on the prior. Then, the rollout policy is employed to simulate the online sampling process several times. Following this, we collect crucial data such as the state information (input) and $Q_t$ (target), which are calculated by the rollout policy after allocating the $t$-th simulation budget. These become the training data for the NN training in the next round.

In the training data generation process, the base policy of the rollout policy is the NN from the previous training round. Since Proposition \ref{pro1} illustrates that the rollout policy has the property of performing at least as well as directly using its base policy with a certain probability, the rollout policy in the current round is expected to be more effective than directly using the NN trained in the previous round for allocation decisions at each step. Consequently, the NN trained using the data generated by the rollout policy in the current round is likely to outperform the NN from the previous round. Note that in the first training round, the untrained NN lacks the capability to guide simulation budget allocation.  Therefore, classic R\&S sampling procedures, such as EA, KG, and OCBA, can be used as the base policy in the rollout in the first round.

Upon completing several training rounds, we conduct a comprehensive simulation test to assess the allocation capabilities of the newly trained NN.   In this test, the NN is directly applied as a policy to guide simulation budget allocation. In the context of fixed-budget R\&S problems, the key metric for evaluation is the improvement in PCS achieved by the new NN, as compared to the performance of the previous NN in identical simulation tests. If the PCS obtained from the new NN is greater than that from the old NN, the training parameters are updated; otherwise, the old NN is retained. The action to allocate the $(t+1)$-th simulation observation determined by the NN is
$$a^{NN}_{t+1}=\arg \underset{i=1,\dots,N}{\max}V_t^{(i)},$$
which means that at each step, the alternative with the largest action value calculated by the NN model is selected. 

\begin{figure}[htbp]
    \centering
    \includegraphics[width=1\textwidth]{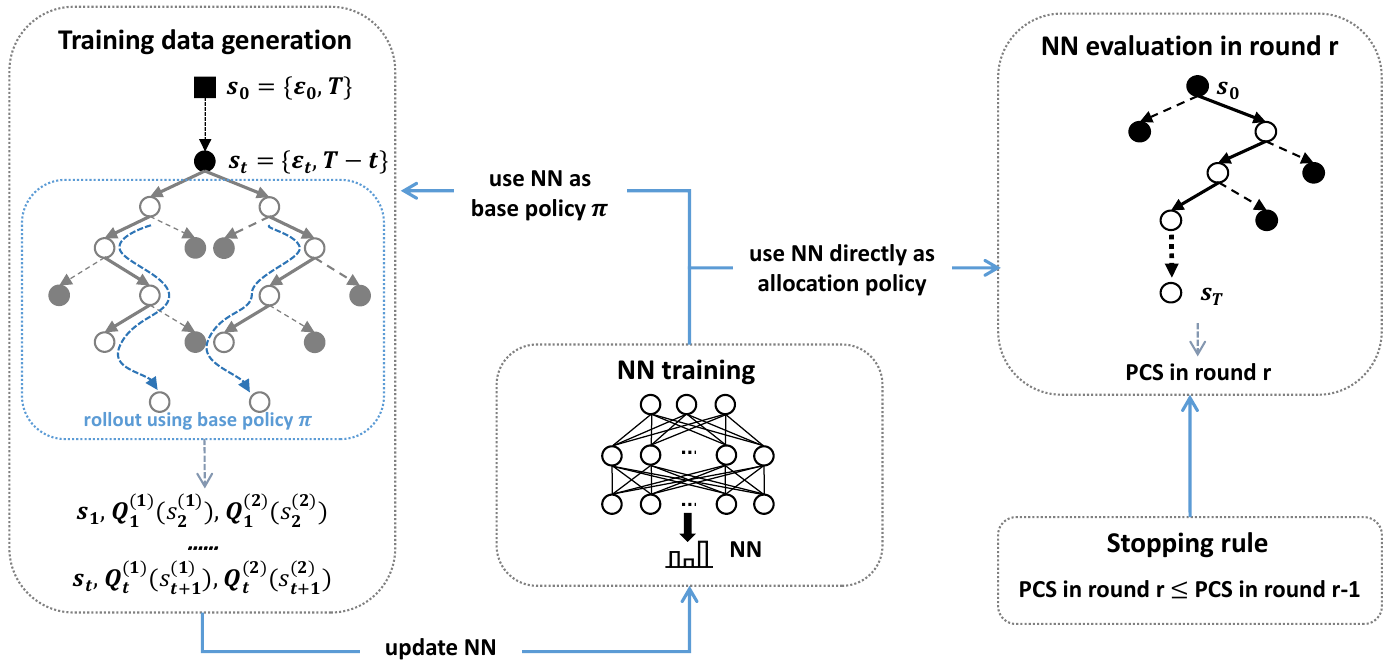}
    \caption{NN training and evaluating architecture with the number of alternatives=2.}
    \label{NN}
\end{figure}

The training process can continue until the NN evaluation does not improve, indicating that the NN has converged, or the training process can be stopped when the PCS improvement is smaller than certain pre-specified level. The pseudo-code of the pre-training process is shown in Algorithm \ref{alg2} and the pipeline of the training is depicted in Figure \ref{NN}. 
\begin{algorithm}  
\caption{Pre-training}  
\LinesNumbered  
\label{alg2}
\KwIn{number of alternatives $N$, times of rollout $K$, number of forward steps in rollout $H$, simulation budget $T$}   
\While{the PCS in evaluation does not satisfy the stopping rule}
{\For{ t=1\  \textbf{to}\  T}  
{   Calculate the current statistics of alternatives $\varepsilon_t$ according to $a_{t}^{roll}$.\\
    \For{k=1\ \textbf{to}\ K} 
    { 
    \For{i=1\ \textbf{to}\ N} 
    {  
     Rollout $H$ steps forward with NN as the base policy and get the reward 
    $r_{t,k}^{(i)}$.
    }  
    }
    Calculate the value function $Q_t^{(i)}=\frac{1}{K}\sum_{k=1}^Kr_{t,k}^{(i)}$.\\
    Collect the training data $\{\varepsilon_t, Q_t\}$.\\
    The sampling action is $a_{t+1}^{roll}=\arg\max_{i=1,\dots,N}Q_t^{(i)}$.
}
NN training: update parameters of NN by minimizing $Loss$ with the Adam optimizer.\\
NN evaluation: NN is used as allocation policy in simulation and then the corresponding PCS is obtained.}
\KwOut{the trained NN} 
\end{algorithm} 
\subsection{Complexity Analysis}\label{section42}
As described in Section \ref{section41}, the pre-training process is divided into two parts: training data generation (rollout) and NN training (parameter updating). The following proposition summarizes the computational complexity of both training the NN and employing it directly as an allocation policy in R\&S problems.
 \begin{proposition}\label{pro_complex2}
     Suppose the number of alternative statistics is $C$, the width of the hidden layer of NN is $W$, the number of hidden layers is $n_h$, the number of alternatives is $N$, and the simulation budget is $T$. Then the computational complexity of NN training is $O((CN+1)W+n_{h}W^2+NW)$. The computational complexity of using the NN as the allocation policy in the R\&S problem is $O(NT)$.
 \end{proposition}
 
\begin{figure}[htb]
\begin{center}
\includegraphics[width=0.5\textwidth]{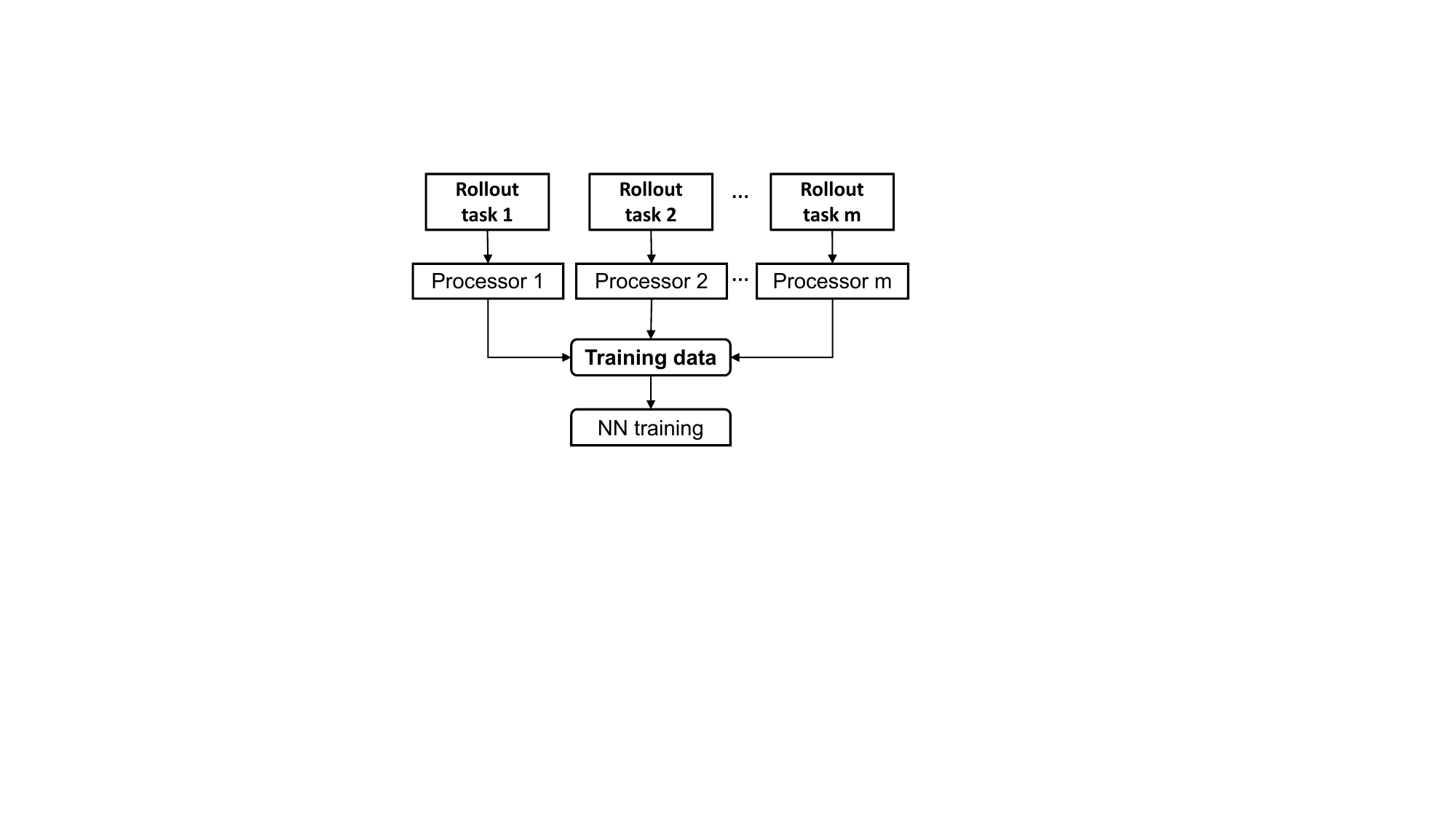}
\caption{The training procedure of NN in parallel.} \label{parallel}
\end{center}
\end{figure}

   From Proposition \ref{pro_complex1}, we know the rollout requires a large amount of computation since the complexity of rollout policy is $O(N^2)$. And from Proposition \ref{pro_complex2} we know the complexity of NN training is $O(N)$. Consequently, within the pre-training process, generating training data using the rollout policy typically demands more time compared to the actual NN training, due to the higher computational complexity of the rollout. The training efficiency is largely affected by the process of generating training data by the rollout policy, because each round of pre-training involves executing multiple rollout tasks to generate a substantial amount of training data, where a rollout task refers to allocating a predefined number of budgets using the rollout policy after getting a problem from the prior. Since each rollout task represents an independent simulation without the need for information synchronization and communication, deploying the rollout tasks across multiple parallel processors can significantly expedite the collection of training data. The diagram of parallel training program using $m$ processors for the NN is shown in Figure \ref{parallel}.

\section{DCR Framework for Large-scale R\&S Problems}\label{section5}
The pre-training process, involving multi-stage iterations, incurs significant computational expense due to the repeated execution of complex operations at each stage. While the parallel training method can expedite the generation of training data, it does not inherently reduce the overall computational complexity, which limits its effectiveness primarily to small or medium-scale problems. To solve this issue, in this section, we first introduce the DCR framework for the fixed-budget R\&S problem, which combines the idea of ``divide-and-conquer" and ``recursion", then we recalculate the complexity of above methods under this framework. 

\subsection{Framework}\label{section51}
Divide-and-conquer involves breaking down a large, complex problem into smaller, more manageable subproblems.   These subproblems are solved individually, with the aggregate solutions constituting the solution to the original problem. The subproblems, resembling smaller versions of the original problem, lend themselves well to the use of recursion. By repeatedly applying divide-and-conquer in a recursive manner, the size of each subproblem is progressively reduced until it becomes straightforward to solve. Furthermore, this framework exhibits significant advantages in a parallel computing environment.  Within such an environment, the task of selecting from different groups can be concurrently executed on multiple processors.  Consequently, there is no need for inter-processor communication until each processor independently determines its local best alternative.  This framework effectively enhances computational efficiency, particularly suited for optimizing resources when addressing large-scale problems.

\begin{figure}[htb]
\begin{center}
\includegraphics[width=0.6\textwidth]{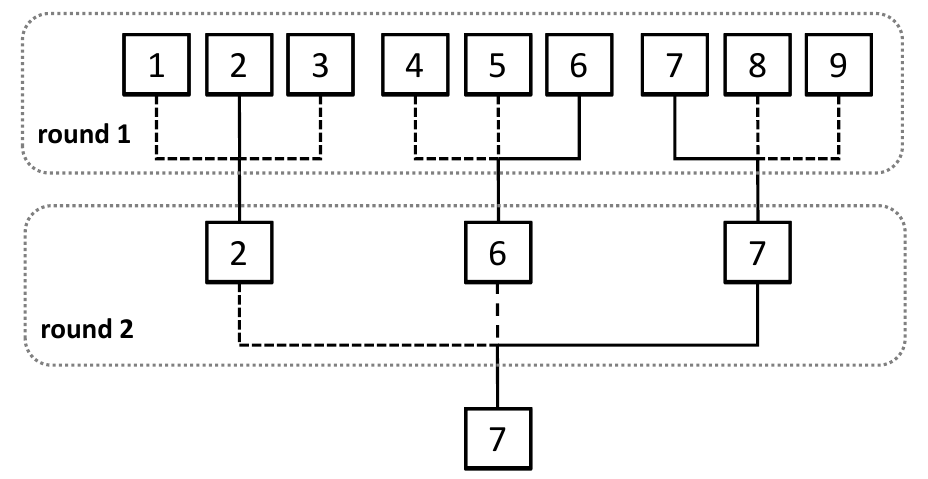}
\caption{The diagram of the DCR framework.} \label{task}
\end{center}
\end{figure}

To be specific, we set the number of the alternatives for the R\&S problem as $N$ and the number of the divided subproblems as $M$, then the best can be selected by at most $\lceil\log_M^N \rceil$ rounds, where  $\lceil\cdot \rceil$ is the ceiling function. In round $r$, $\frac{N}{M^{r-1}}$ alternatives are divided into $\lceil\log_M^N \rceil-r+1$ groups of size $M$, where $r=1,2,\dots,\lceil\log_M^N \rceil$. Each group of alternatives is screened using an R\&S procedure. The selected alternatives will advance to the next round of comparison, while the rest will be eliminated. This process of grouping and screening repeats until only a single alternative is remaining. About $\frac{M-1}{M^r}N$ alternatives can be eliminated in round $r$. Figure \ref{task} is the DCR framework diagram of recursively solving the problem with 9 alternatives by dividing it into subproblems with 3 alternatives.

Within the DCR framework, the selection process is simplified such that the best alternative is chosen within its respective group during each round, and the PCS under the DCR framework is no longer the same as (\ref{eqpcs}). Let the index of the best alternative be $b$, and $\mu_b^{true}=\arg\max_{i=1\dots,N}\mu_i^{true}$. In each round, we only focus on the screening results of the group, to which the alternative $b$ belongs. Then, the PCS in round $r$ is
 \begin{equation*}
    \text{PCS}_r=\text{Pr}(\text{alternative $b$ is selected in round $r$}),
 \end{equation*}
where the selection in the group of round $r$ is based on (\ref{eq2122}) or (\ref{eqselect1}), depending on the distribution assumption and objective. Within the DCR framework, the optimal alternative no longer needs to be compared against all remaining $N-1$ alternatives.  Instead, it requires comparison with at most $\lceil\log_M^N \rceil M$ alternatives, significantly reducing the likelihood of erroneous exclusion.  The relationship between the $\text{PCS}_r$ of the optimal alternative within its group and the overall PCS in each round can be established using the Bonferroni’s inequality, giving a lower bound for the PCS, i.e., the final PCS for the fixed-budget R\&S problem satisfies
\begin{align}
    \text{PCS}&=\text{Pr}\left(\bigcap_{r=1}^{\lceil\log_M^N \rceil}\text{\{alternative $b$ is selected in round $r$\}}\right)\notag\\
    &\geq 1-\sum_{r=1}^{\lceil\log_M^N \rceil}\text{Pr}\left(\text{alternative $b$ is eliminated in round $r$}\right)\notag\\
    &=1-\sum_{r=1}^{\lceil\log_M^N \rceil}\left(1-\text{Pr}\left(\text{alternative $b$ is selected in round $r$}\right)\right)\notag\\
    &=\sum_{r=1}^{\lceil\log_M^N \rceil}\text{PCS}_r-\lceil\log_M^N \rceil+1.\label{proofpro11}
\end{align}

\cite{Hong2022SolvingLF} introduce the FBKT procedure, which is a special case using EA for sample-allocation in the DCR framework, and each group in round $r$ is equally allocated the same number of simulation observations, i.e., each alternative in round $r$ can be sampled $\frac{M^{r-1}}{N}T_r$ times, where $T_r$ is the total simulation budget of round $r$. For fixed-budget  R\&S problems, simulation budget $T$ needs to be allocated to each round in advance, such that $\sum_{r=1}^{\lceil\log_M^N \rceil}T_r\leq T$. Then $T_r$ simulation observations are allocated to each group in round $r$. Allocation under the DCR framework involves the issue of how much budget to allocate in each round. From the frequentist perspective, \cite{Hong2022SolvingLF} prove that when the budget $T_r$ in round $r$ satisfies that 
\begin{gather}\label{trallocation}
    T_r=\frac{r}{\phi(\phi-1)}\left(\frac{\phi-1}{\phi}\right)^rT,
\end{gather} for $r\geq 1$ and $\phi\geq 2$, FBKT attains the rate for maintaining correct selection (RMCS) in the order of $N$, which is a growth rate of simulation budget $T$ to guarantee that as $N$ increases, the PCS does not decrease to zero. In the DCR framework, the scale and number of groups into which the remaining alternatives are divided in each round remain uncertain (we fixed the group size as $M$ above for better interpretation and analysis). This division is contingent upon the scale of the original problem. Furthermore, the flexibility to employ any R\&S algorithm for intra-group filtering complicates the establishment of a definitive lower bound for PCS, as the effectiveness of these algorithms can vary. While the results presented here are derived within a Bayesian context, they have broader applicability, aligning well with those obtained under the frequentist framework, particularly in terms of reducing the requisite comparisons for identifying the optimal solution.  An intriguing direction for future theoretical research lies in deriving conclusions within the Bayesian framework that parallels the RMCS concept, potentially offering novel perspectives and methodologies in large-scale R\&S problems.
 
The problem of how to divide groups is also faced in the DCR framework. Instead of randomly grouping alternatives, we can also use the seeding method developed by \cite{Hong2022SolvingLF}, analogous to that the seeded players in the ball game are allocated to different groups.
\subsection{Complexity Analysis in DCR Framework}\label{section52}
After obtaining a small-scale and high-precision NN model, we can use it in the DCR framework proposed in Section \ref{section41} to solve large-scale R\&S problems. In this subsection, for the purpose of facilitating theoretical analysis, we continue to assume the use of a single-scale NN model for intra-group filtering. When we have the NN model of scale $M$, problems of scale $N$ under the same distributional assumptions can be solved within $\lceil\log_M^N \rceil$ rounds under the DCR framework, wherein the computational complexity exhibits a logarithmic increase relative to $N$, significantly enhancing computational efficiency as detailed in Corollary \ref{coro_complex1}.
\begin{corollary}\label{coro_complex1}
When a pre-trained NN model for $M$ alternatives is used to guide the allocation policy to solve the problem with $N$ alternatives under the DCR framework, the computational complexities of rollout policy and NN training in the presence of the number of alternatives are $O(M^2)$ and $O(M)$, respectively. The computational complexity of diectly using NN as the allocation policy is $O(\lceil\log_M^N \rceil)$.
\end{corollary}
\begin{remark}
In the absence of the DCR framework, the computational complexities of the rollout policy and NN training with respect to the number of alternatives, $N$, are $O(N^2)$ and $O(N)$, respectively. When $M$ is substantially smaller than $N$, training the NN within the DCR framework leads to a significant reduction in computational resources. Furthermore, using an NN of scale $N$ to address a problem of similar scale results in a computational complexity of $O(N)$. Under the DCR framework, the complexity for solving the same problem is reduced to $O(\lceil\log_M^N \rceil)$, transitioning from linear to logarithmic scale, thereby increasing solution efficiency.
\end{remark}
\section{AlphaRank Algorithm}\label{section6}
So far, all the building blocks of our approach have been introduced. In this section, we give a formal description of AlphaRank algorithm. AlphaRank algorithm is designed to pre-train a series of high-precision NN models under a given prior.  Subsequently, these trained NN models can be directly employed for allocation decisions in practical problems.

The pre-training process of AlphaRank is carried out offline based on a given prior distribution, which is described in detail in Section \ref{section4}. During the pre-training, the rollout policy is adopted for several rounds to improve the ablity of the NN. All the data used in the pre-training is sampled based on the prior. Given the unknown scale of practical problems, we train NN models for R\&S problems of varying scales, ranging from 2 to 10, tailored to a specific prior.  During training, we set a maximum forward rollout step count, denoted as $H$.  This implies that if we encounter a scenario where the actual remaining budget exceeds $H$, the action values we obtain are short-sighted approximations of the true values.

We utilize the trained NN models for online sampling decisions. When the scale of the practical problem does not exceed the maximum scale of our pre-trained models, we directly use the corresponding NN for sampling decisions.  Otherwise, in conjunction with the DCR framework proposed by Section \ref{section5}, we first decompose the problem into smaller subproblems that can be directly solved with our pre-trained NN models, and then recursively address them. Note that in practice, it is not necessary to ensure that the number of alternatives within the group is the same in each round and between rounds, as long as these subproblems can find a suitable NN model to solve.  When the remaining budget in the practical problem is greater than the maximum step count $H$ predefined in training, the remaining budget part of the model input should consistently be $H$ until the actual remaining budget falls below $H$.

It is important to note that the pre-training process is a standalone and one-time procedure.  Once the NN models are trained on the specific prior, the resultant NN models can be repeatedly applied to a multitude of practical problems with the same prior.  This process is akin to AlphaGo --- once trained and released online, it becomes accessible to various users.  We can develop a standardized method for pre-training these models and publish them on cloud platforms like GitHub, thereby making them widely available and eliminating the need for repetitive training.
\section{Numerical Experimental Results}\label{section7}
In this section, we examine the performance of the proposed procedures under different settings. The objectives of the numerical experiments are threefold: (1) to compare the performances of our rollout policy and some existing procedures;  (2) to show the AlphaRank procedure can indeed train NN model that can capture the behavior of rollout policy, and greatly improve the performance compared to the existing procedures;  (3) to illustrate training small-scale models and combining them with the DCR framework to solve large-scale problems can not only improve training efficiency but also the statistical performance.

All experiments use the selection policy with the largest posterior mean as the selection policy. The comparison results between the selection policy with the largest posterior PCS and the selection policy with the largest posterior mean are presented in Appendix \ref{APPD:SUPP0}. The rationale behind opting for the latter as our primary selection policy is grounded in a balance between effectiveness and computational efficiency.   While the former policy shows greater effectiveness in low-confidence scenarios, this advantage does not sufficiently offset the increased computational demands, particularly in high-confidence scenarios where the superiority of the optimal selection policy becomes marginal. All performance statistics in this section are estimated from $10^5$ independent macro-simulations with 100 rollouts per step. All codes used in this section can be found in \href{https://github.com/Hannah-Zhou/AlphaRank-General-AI-Framework-for-Ranking-and-Selection}{AlphaRank.github}.
\subsection{Numerical Experiment of Rollout Policy}\label{section71}
In this experiment, the number of alternatives is $N = 10$. Parameter settings can be categorized as a high-confidence scenario and a low-confidence scenario considered in \cite{Peng2018b}. The following four sample-allocation policies are used for comparison: EA that sequentially allocates the alternatives from the first to the last in a cyclical manner; KG with uninformative prior (\citealt{2007A}); AOAP with uninformative prior (\citealt{Peng2018b}); the “most
starving” sequential OCBA (\citealt{chen2011}); and the rollout policy proposed in Section \ref{section3}. 

\textbf{Example 1 (high-confidence):} High-confidence scenario can be qualitatively described by three characteristics: large differences between the means of competing alternatives, small variances, and large simulation budget (\citealt{Peng2018b}). We consider a setting that the total simulation budget is $T=400$, $\mu_i^{(0)}=0$, $\sigma_i^{(0)}=1$, and the true variance is $(\sigma_i^{true})^2=1$, $i=1,2,\dots,10$. 

\textbf{Example 2 (low-confidence):} Low-confidence scenario has characteristics that the differences between the means of competing alternatives are small, the variances are large, and the simulation budget is small (\citealt{Peng2018b}). We consider a setting that the total simulation budget is $T=200$, $\mu_i^{(0)}=0$, $\sigma_i^{(0)}=0.02$ for $i=1$, $\sigma_i^{(0)}=0.01$ for $i=2,3,\dots,10$, and the true variance is $(\sigma_i^{true})^2=\sqrt{5}$, $i=1,2,\dots,10$. 

The initial 100 simulation observations are allocated equally to each alternative, and the remaining simulation observations are allocated according to different sample-allocation procedures. 
\begin{figure}[htbp]
    \centering
    \subfigure[High-confidence scenario]{
        \includegraphics[width=2in]{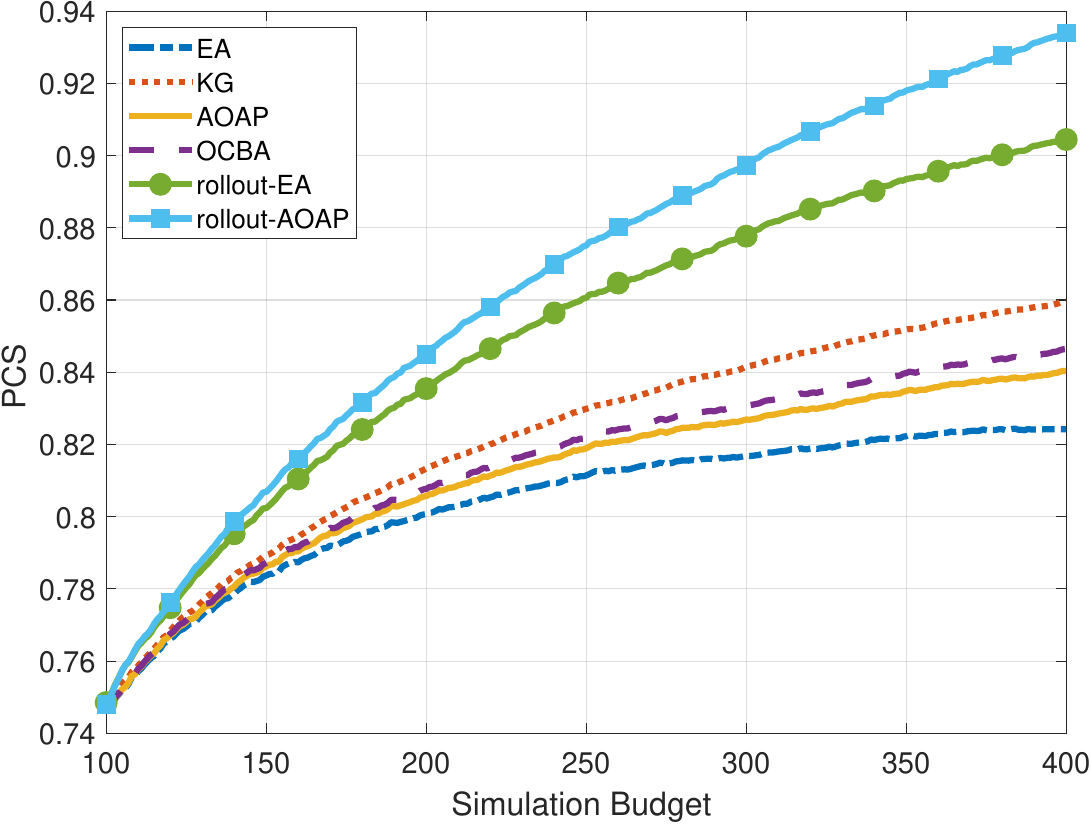}
    }
    \subfigure[Low-confidence scenario]{
	\includegraphics[width=2in]{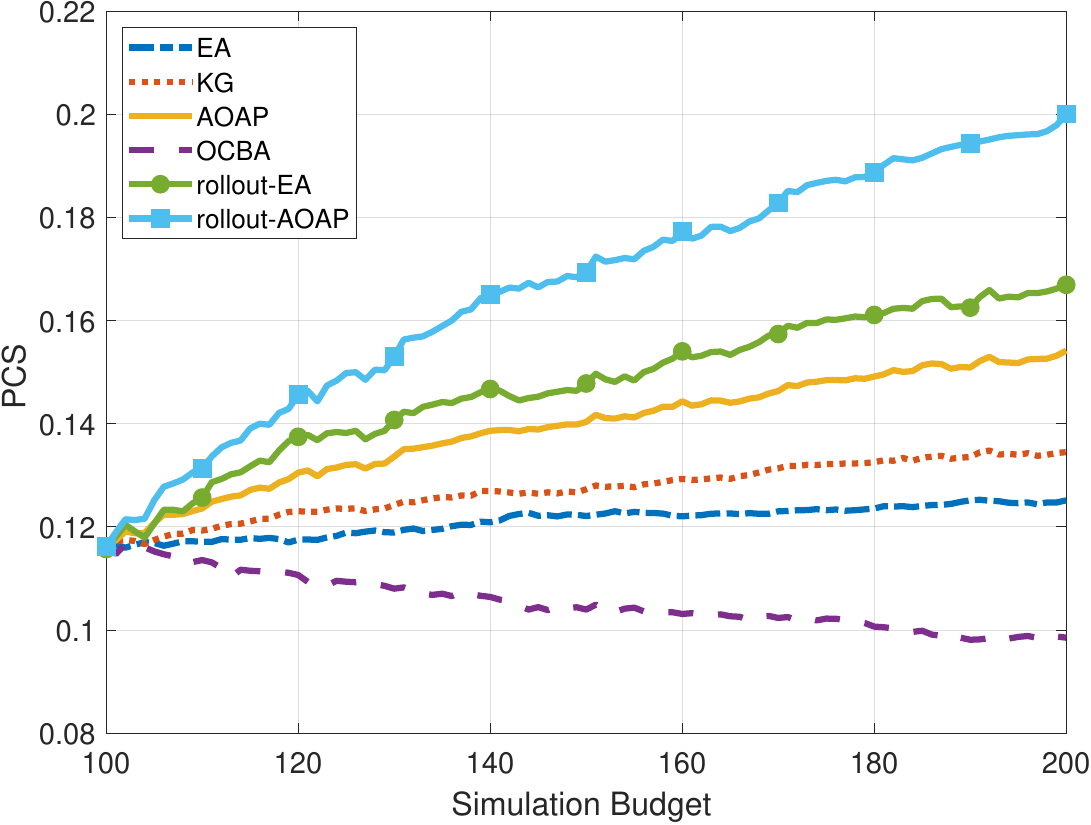}
    }   
    \caption{PCSs of EA, KG, AOAP, OCBA and rollout policies.}
    \label{roll-normal}
\end{figure}
 
Figures \ref{roll-normal} (a) and \ref{roll-normal} (b) illustrate the progression of PCS as a function of the simulation budget for each scenario, providing a visual representation of the comparative effectiveness of the different policies. The experimental results demonstrate that, as the simulation budget increases, rollout-EA and rollout-AOAP, whose base policies are EA and AOAP, respectively, outperform other methods in both examples. Specifically, in the high-confidence scenario, the PCSs of KG, OCBA, AOAP, and EA grow steadily and reach a range between 82\% and 86\%, whereas the PCSs achieved by rollout-EA and rollout-AOAP grow significantly faster than those obtained by other methods. In the low-confidence scenario, the PCSs of the four classic sequential policies hardly increase or even decrease with the increase of the simulation budget, and the PCSs of rollout-EA and rollout-AOAP increase at a faster rate than others. A key observation is that the efficacy of the rollout policy is enhanced when its base policy (EA or AOAP) demonstrates stronger performance, suggesting a synergistic relationship between the base and rollout policies. To reach the PCS level achieved by AOAP and EA when the simulation budget is exhausted, rollout-EA and rollout-AOAP save about 57\% and 45\% of the simulation budget respectively in the high-confidence scenario, and 97\% and 38\% of the simulation budget respectively in the low-confidence scenario.

\subsection{Numerical Experiment of AlphaRank}\label{section72}
To evaluate the effectiveness of AlphaRank, we conduct performance tests on the model, which has been pre-trained as per the methods outlined in this paper and using EA, KG, or AOAP as the rollout base policy for generating the first round of training data. We compare it with EA, KG, AOAP. In this experiment, we provide two small-scale problems in high-confidence scenario (Example 1) and low-confidence scenario (Example 2), and a large-scale problem (Example 3) to apply the DCR framework. 

\textbf{Example 1 (high-confidence):}  The number of alternatives is $N = 10$. The total simulation budget is $T=200$, $\mu_i^{(0)}=0$, $\sigma_i^{(0)}=1$, and the true variance is $(\sigma_i^{true})^2=1$, $i=1,2,\dots,10$. 

\textbf{Example 2 (low-confidence):} $N = 10$, $T=200$, $\mu_i^{(0)}=0$, $\sigma_i^{(0)}=0.02$ for $i=1$, and $\sigma_i^{(0)}=0.01$ for $i=2,\dots,10$, and the true variance is $(\sigma_i^{true})^2=1$, $i=1,2,\dots,10$.

We allocate 100 initial simulation observations to each of the 10 alternatives for the first stage, with the remaining 100 simulation observations allocated using the various sample-allocation procedures. 
\begin{figure}[htbp]
    \centering
    \subfigure[High-confidence scenario]{
        \includegraphics[width=2in]{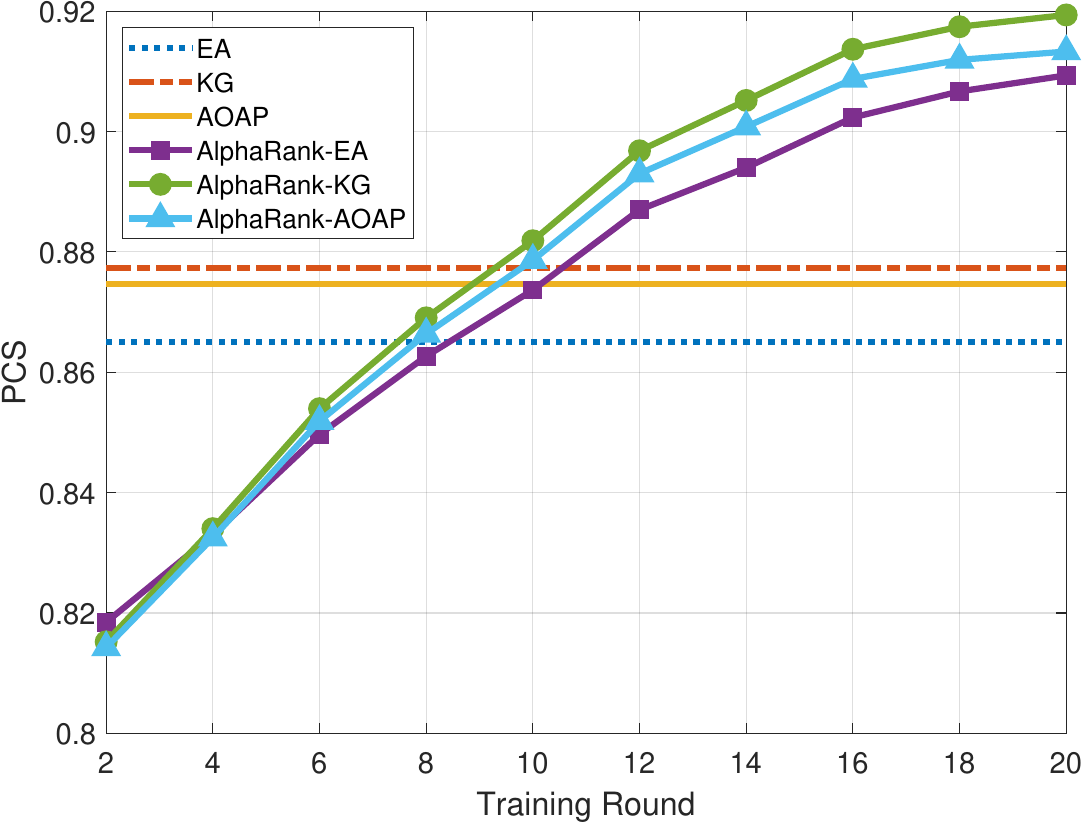}
    }
    \subfigure[Low-confidence scenario]{
	\includegraphics[width=2in]{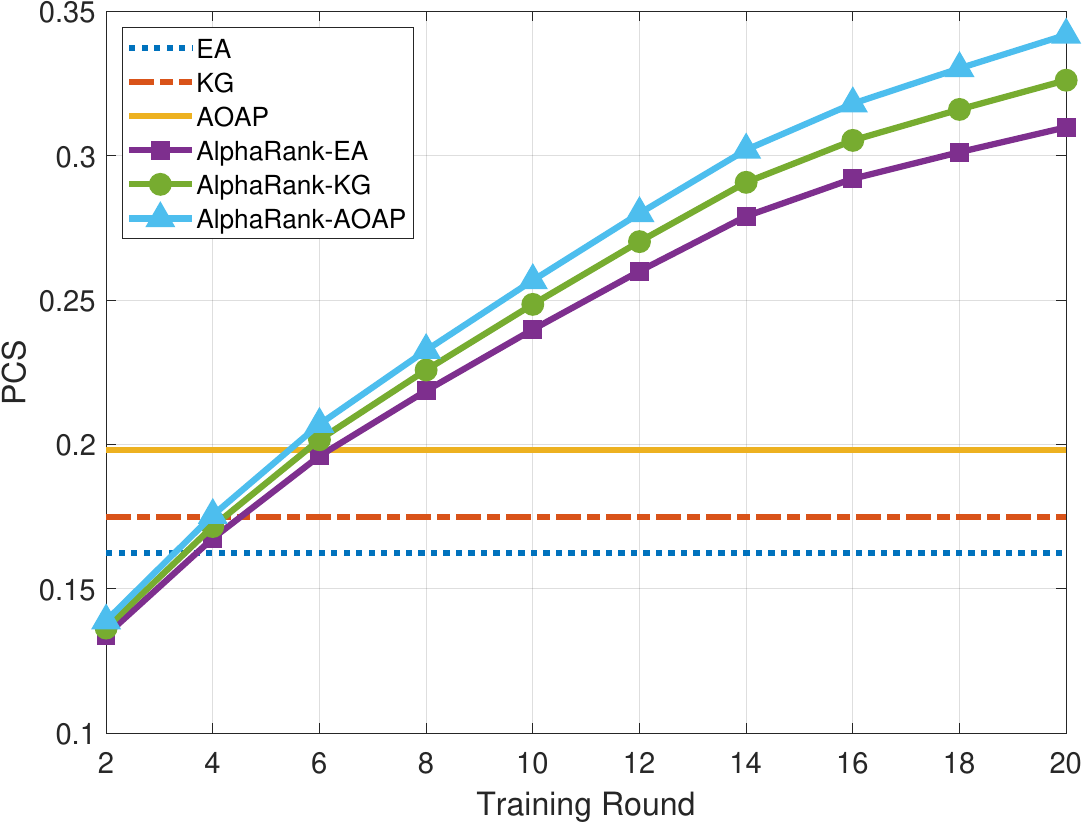}
    }
   
    \caption{PCSs of EA, KG, AOAP, and AlphaRank in Examples 1 and 2.}
    \label{normal}
\end{figure}

 As shown in Figure \ref{normal}, the PCSs of AlphaRank-EA, AlphaRank-KG, and AlphaRank-AOAP, whose base policies are EA, KG, and AOAP for training NN in the first round, respectively,  gradually increase with the number of rounds and eventually exceed the PCSs of EA, KG, and AOAP. The gaps between AlphaRank and benchmarks increase as the training round increases demonstrates the effectiveness of the design of the pre-training process, with the performance of the NN improving as the number of training rounds increases. Significantly, AlphaRank begins to demonstrate a notable advantage over traditional R\&S methods after 10 and 4 rounds in the high-confidence and low-confidence scenarios, respectively. With EA as a baseline, the performance of AlphaRank improves by 4\% and 20\% in both examples. These numerical experiments suggest that AlphaRank can improve the performance of base policies, especially in scenarios where the base policies perform poorly.
 
\textbf{Example 3 (large-scale):} $N = 10000$, $T=220000$, $\mu_i^{(0)}=0$, $\sigma_i^{(0)}=1$, and $(\sigma_i^{true})^2=1$, $i=1,2,\dots,10000$. 

To address the large-scale R\&S problem, we employ the DCR framework, comparing AlphaRank's performance when using models of scales M=10 (AR10) and M=100 (AR100). AR10-EA, AR10-KG, and AR10-AOAP are three NN models with EA, KG, and AOAP as base policies respectively for the first round training in Example 1. In this case, the numbers of rounds of screening are 4 and 2 when using AR10 and AR100, receptively. We follow the practice of \cite{Hong2022SolvingLF} and allocate the budget in each round based on (\ref{trallocation}), and we use $\phi=2$ in this example. AR100-EA, AR100-KG, and AR100-AOAP are trained in parallel using 32CPUs for the setting of Example 3. Numerical results in other settings can be found in Appendix \ref{APPD:SUPP3}.

\begin{table}[htb]
\centering
\small
\renewcommand{\arraystretch}{1.5}
\caption{\label{tab:sub2}Training time, simulation time, and PCSs of AlphaRank with different scales in Example 3.}
\begin{tabular}{c|cccccc}
\hline
Metric&	AR10-EA&	AR100-EA&	AR10-KG&	AR100-KG&	AR10-AOAP&	AR100-AOAP\\\hline
\makecell[c]{CPU Time\\ for 1 Training Round}&	0.71h&	21.27h&	0.78h&	22.18h&	1.08h&	28.16h\\\hline
\makecell[c]{CPU Time\\ for 1 Simulation}&	38.42s&	203.74s&	38.9s&	204.11s&	40.86s&	204.48s\\\hline
PCS in Example 3&	0.8203&	0.8091&	0.8267&	0.8165&	0.8312&	0.8164\\\hline
\end{tabular}
\end{table}
As detailed in Table \ref{tab:sub2}, the CPU time required for a single training round of AR10 and AR100 (including data generation and NN parameter updating) reveals that AR100's training is substantially more time-intensive.     However, despite the greater computational resource investment, AR100's final PCS is even lower than that of AR10, underscoring the efficacy of training high-precision, small models for large-scale R\&S problems under the DCR framework, which not only greatly reduces the training time but also improves the actual PCS. In large-scale R\&S problems, using the DCR framework to eliminate poor-performing alternatives in advance can greatly improve learning efficiency. Appendix \ref{APPD:SUPP4} presents numerical results on PCS and EOC for the policies that apply the DCR framework and those that do not.

\section{New Insights Revealed by AlphaRank}\label{section8}
In this section, we want to explore potential reasons behind its superior performance by analyzing AlphaRank's behavior through numerical experiments.  The problems to be investigated in this numerical studies are threefold: (1) the reason why AlphaRank has good performance in low confidence scenarios where most algorithms do not perform well; (2) the differences in the behavior of various procedures under different confidence scenarios; (3) the change of sampling behavior of various procedures with the increase of problem scale.
\subsection{Traditional Procedures}\label{section81}
Traditionally, the design of effective sampling policies is often guided by the mean-variance trade-off, where the goal is to balance the allocation of simulation observations between alternatives with better mean performance and those with larger variances. For example, in the classic R\&S setting where $X_{i,t}\sim N(\mu_i^{true}, (\sigma_i^{true})^2)$, the OCBA method considers the following static optimization problem (\citealt{2000Simulation}):
\begin{equation}  \label{eqbest}  \max_{T_1+\dots+T_N=T}\text{Pr}\left(\Bar{X}_b\geq\max_{i=1,\dots,N}\Bar{X}_i\right),
\end{equation}
where $\Bar{X}_i$ is the sample mean obtained after allocating $T_i$ simulation observations to alternative $i$, and $b$ is the index of the best
alternative. The sampling ratio $r_i$ of each alternative is calculated by $r_i= T_i/T$. To simplify the computation of (\ref{eqbest}), \cite{1371364} derive the analytical form of sampling ratio $r_i^*$ optimizing the large deviations rate of PICS:
\begin{gather}\label{rate1}
  \frac{(\mu_i^{true}-\mu_b^{true})}{(\sigma_i^{true})^2/r^*_i+(\sigma_b^{true})^2/r^*_b}=\frac{(\mu_j^{true}-\mu_b^{true})}{(\sigma_j^{true})^2/r^*_j+(\sigma_b^{true})^2/r^*_b},
\end{gather}
\begin{gather}\label{rate2}
r^*_b=\sigma_b^{true}\sqrt{\sum_{i\neq b}\frac{(r^*_i)^2}{(\sigma_i^{true})^2}},
\end{gather}
 where $i,j\neq b$, $\sum_{i=1,\dots,N}r^*_i=1$. With true parameters plugged in, a static optimal policy (SOP) can allocate simulation observations with the sampling ratio determined by (\ref{rate1}) and (\ref{rate2}). The sampling ratio of SOP is approximately equal to the theoretical sampling ratio of OCBA, which clearly reflects the mean-variance trade-off. Furthermore, many sequential sampling policies are derived to achieve this static sampling ratio asymptotically. However, recent studies, including \cite{7408526} and \cite{Peng2018b}, have revealed that in low-confidence scenarios, adhering strictly to the mean-variance trade-off can result in misleading outcomes, particularly due to the overlooked aspect of induced correlation. In the next few experiments, we compare the sampling behavior of AlphaRank, SOP, and sequential OCBA with the ``most starving'' rule (\citealt{chen2011}) to see that AlphaRank may capture induced correlations that have been overlooked by most traditional methods.

 \subsection{Sampling Behavior in Low-Confidence Scenarios}\label{section82}
In this experiment, the total budget is $T=60$. The initial 15 simulation observations are allocated equally among three alternatives. In the first example, the differences among the alternatives lie on their means, whereas in the second example, the variances of the three alternatives are significantly different. 
 
\begin{figure}[htbp]
    \centering
        \subfigure[Sampling ratio curves of alt. 1]{
        \includegraphics[width=2in]{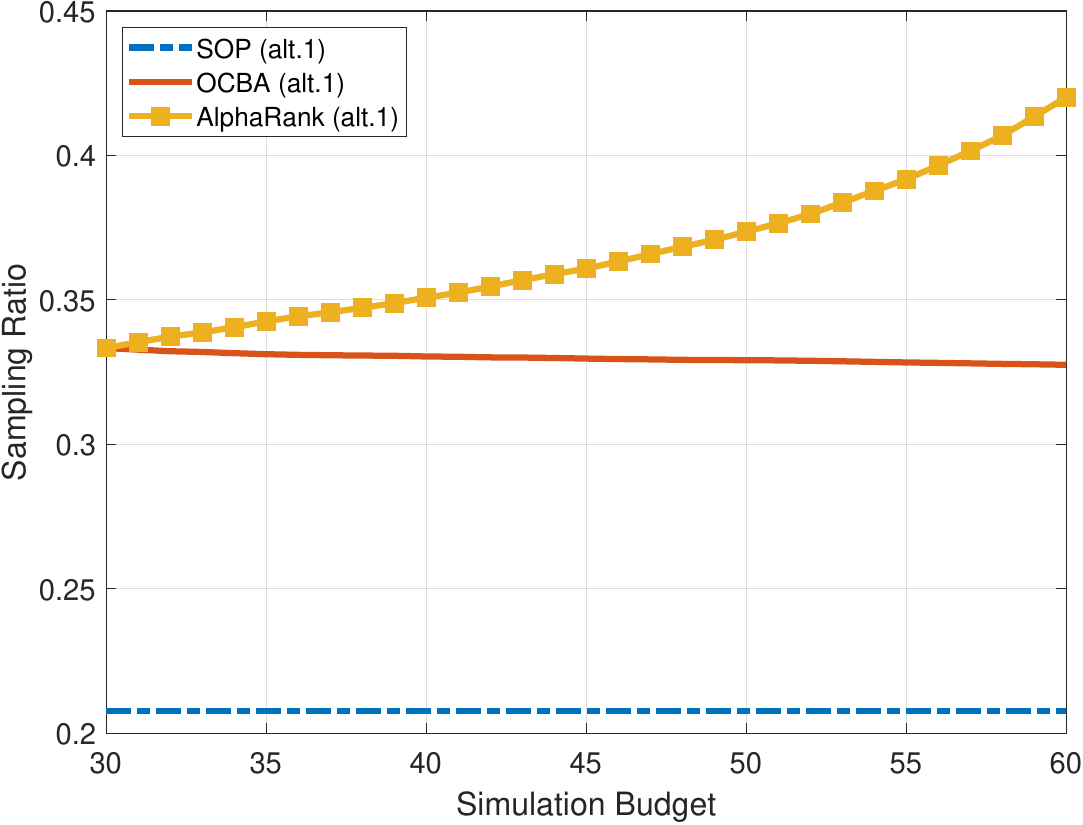}
    }
    \subfigure[Sampling ratio curves of alt. 2]{
	\includegraphics[width=2in]{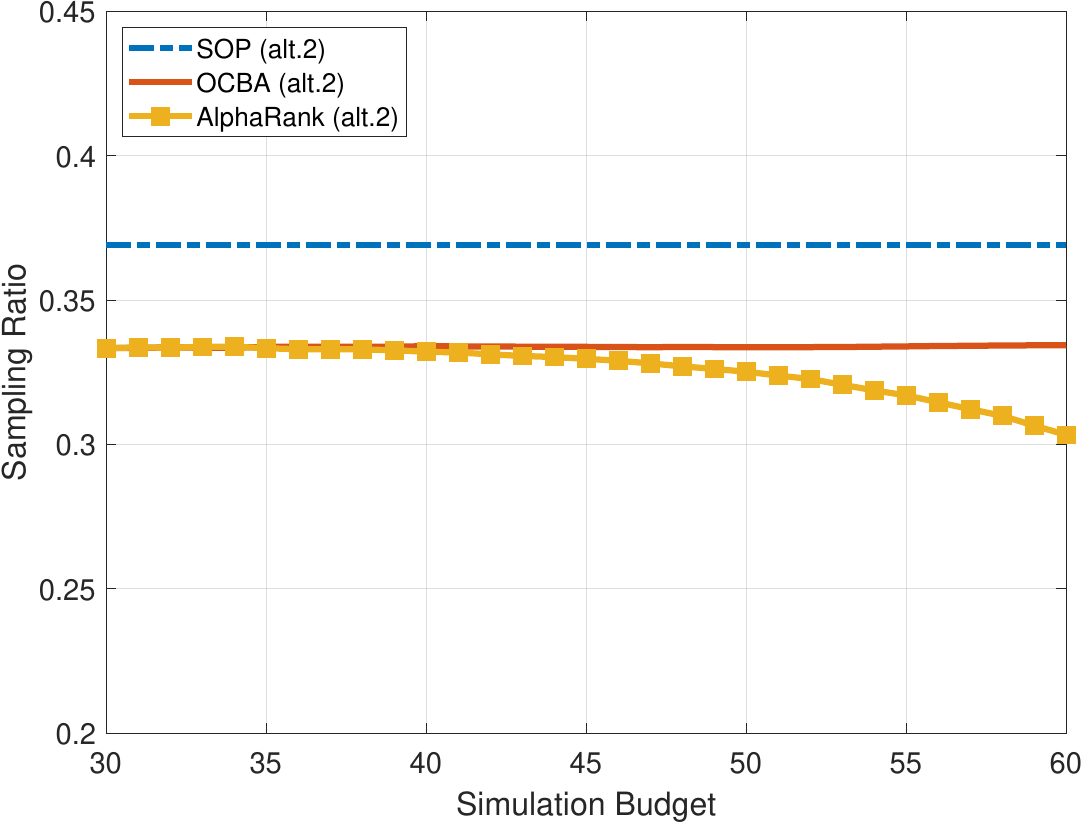}
    }  
\\
    \subfigure[Sampling ratio curves of alt. 3]{
	\includegraphics[width=2in]{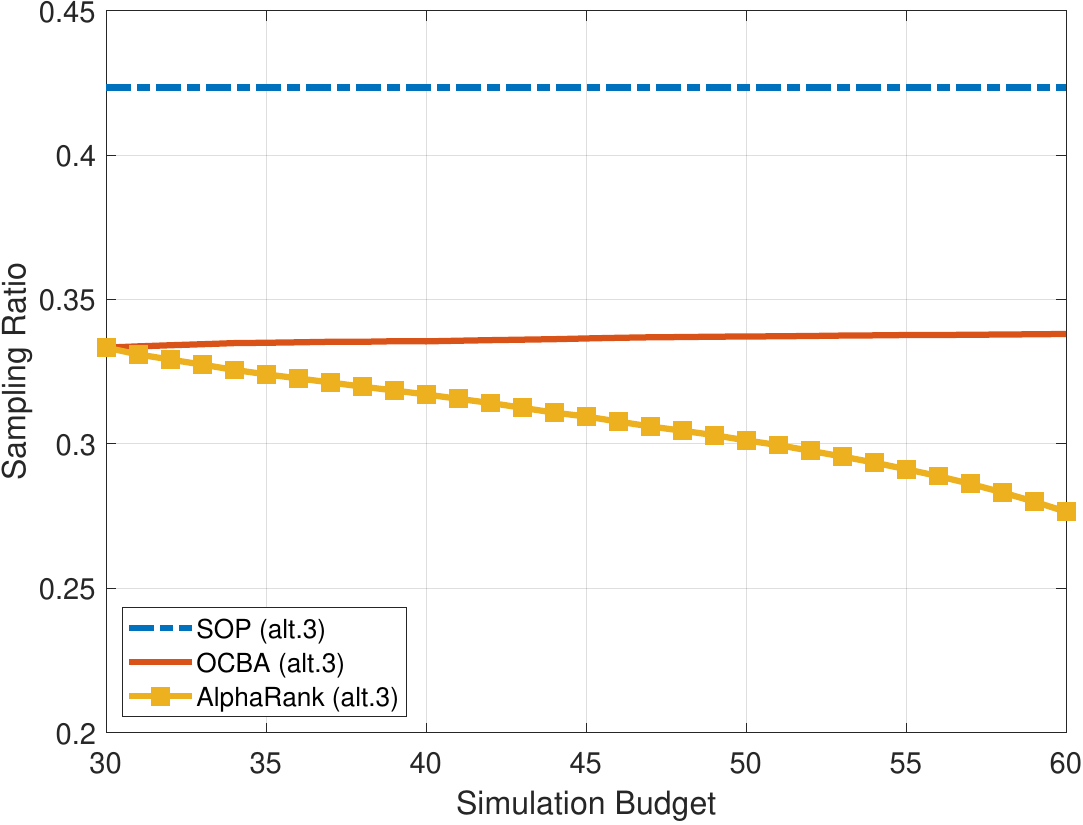}
    }   
        \subfigure[PCS curves]{
	\includegraphics[width=2in]{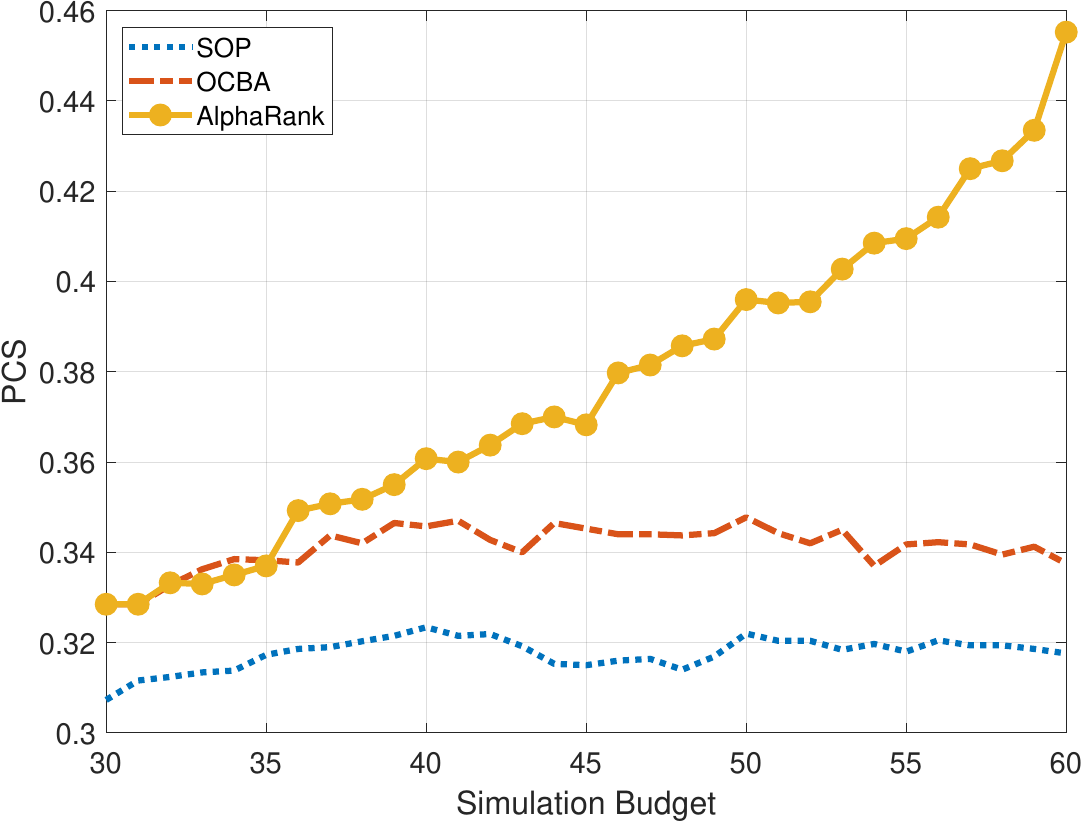}
    }
    \caption{Sampling ratios and PCS curves of SOP, OCBA, and AlphaRank in Example 1.}
    \label{ratio}
\end{figure}

\textbf{Example 1:} $\mu_1^{true}=0.001$, $\mu_2^{true}=0.002$, $\mu_3^{true}=0.005$, and $(\sigma_1^{true})^2=(\sigma_2^{true})^2=(\sigma_3^{true})^2=1$. In Figure \ref{ratio}, we present the sampling ratio and PCS curves of SOP, OCBA, and AlphaRank as the simulation budget grows.

As depicted in Figure \ref{ratio}, SOP exhibits a tendency to allocate more simulation observations to alternatives with higher means.  However, this approach results in PCS performance that is inferior to even random selection (1/3), indicating a potential inefficiency in its allocation strategy. There is a large gap between the sampling ratio of OCBA and the corresponding sampling ratio of SOP, although OCBA still tends to allocate more simulation observations to the alternative with a larger mean as SOP. From Figure \ref{ratio} (a), (b), and (c),  OCBA requires a large amount of simulation budget to asymptotically approach the SOP sampling ratio. Figure \ref{ratio} (d) shows that SOP leads to even worse performance than OCBA, whose PCS slowly decreases with the increase of simulation budget. In this low-confidence scenario, allocating the alternative with better mean performance is likely to decrease the induced correlation and then reduce the PCS. In stark contrast, AlphaRank's strategy involves allocating more simulation observations to alternatives with lower mean performance, diverging significantly from the approaches of SOP and OCBA. This counterintuitive approach explains why AlphaRank's PCS increases, while the PCSs of SOP and OCBA remain relatively stagnant. 

\textbf{Example 2:} $\mu_1^{true}=0$, $\mu_2^{true}=0$, $\mu_3^{true}=0.001$. In the first case, $(\sigma_1^{true})^2=1$, $(\sigma_2^{true})^2=2$ and $(\sigma_3^{true})^2=3$. In the second case, $(\sigma_1^{true})^2=3$, $(\sigma_2^{true})^2=2$ and $(\sigma_3^{true})^2=1$.

 \begin{figure}[htbp]
    \centering
        \subfigure[Case 1]{
        \includegraphics[width=2in]{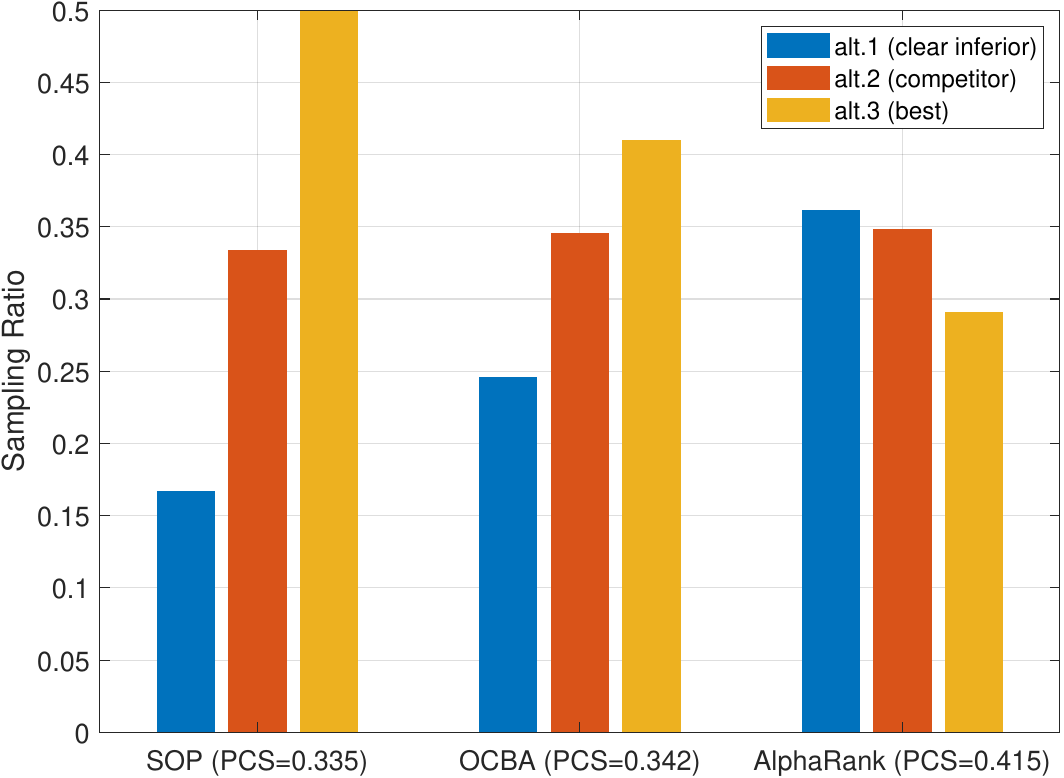}
    }
    \subfigure[Case 2]{
	\includegraphics[width=2in]{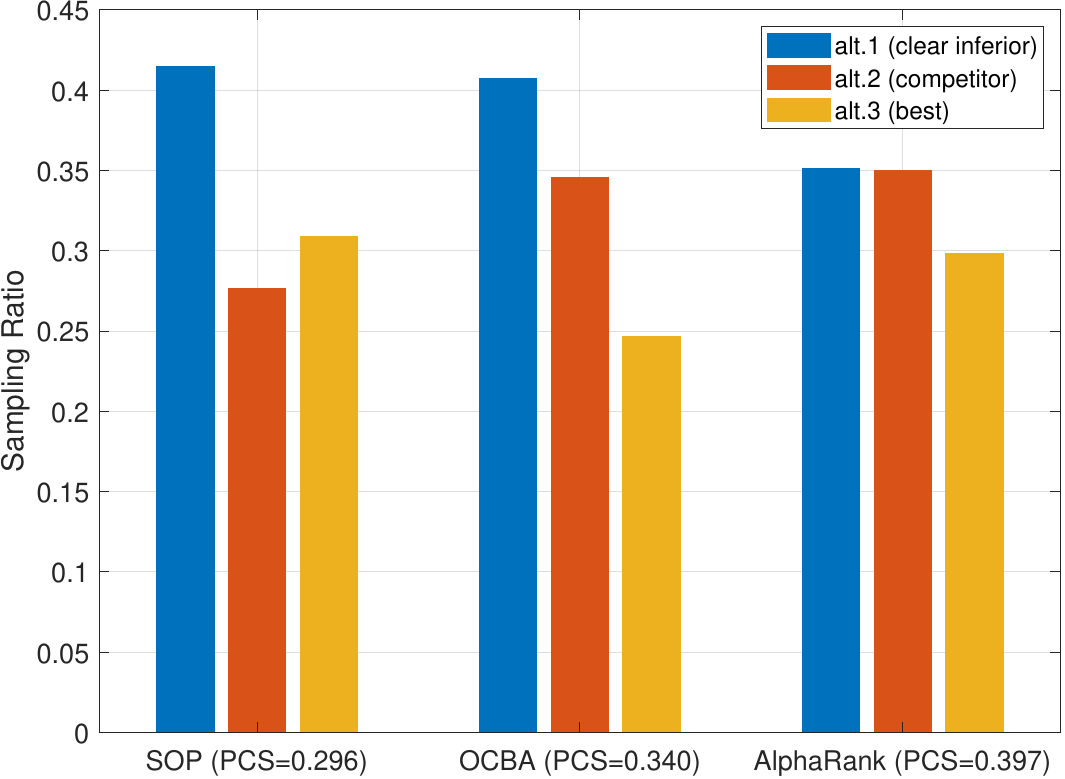}
    }   
    \caption{Sampling ratios of SOP, OCBA, and AlphaRank in Example 2.}
    \label{ratiovar}
\end{figure}

Figure \ref{ratiovar} shows the sampling ratios after 60 simulation observations are allocated following different sampling policies in both cases. Because the differences in means are minimal, both SOP and OCBA allocate more simulation observations to alternatives with larger variances. In Figure \ref{ratiovar}(a), AlphaRank's behavior markedly contrasts with the others by allocating the least simulation observations to the alternative with the largest mean and variance, intelligently avoiding decreasing the induced correlation with a high probability. In Table \ref{tab:exxvar1}, we can see that PCSs of AlphaRank are largest and PCSs of SOP are smallest in all examples.
\begin{table}[htb]
\centering
\small
\caption{\label{tab:exxvar1}PCSs of SOP, OCBA, and AlphaRank.}
\begin{tabular}{c|ccc}
\hline
Scenario	&	SOP	&	OCBA	&	AlphaRank	\\\hline
Example 1	&	0.319	&	0.339	&	0.458	\\
Example 2: Case 1	&	0.335	&	0.342	&	0.415	\\
Example 2: Case 2	&	0.296	&	0.340	&	0.397	\\
\hline
\end{tabular}
\end{table}

\subsection{Sampling Behavior in Different Confidence Scenarios}\label{section83}
The number of alternatives $N$ is 3. The corresponding parameter settings are shown in Table \ref{tab:paras7}, where $\mu^{true}=[\mu_1^{true},\mu_2^{true},\mu_3^{true}]$ and $(\sigma^{true})^2=[(\sigma_1^{true})^2, (\sigma_2^{true})^2, (\sigma_3^{true})^2]$. The initial 15 simulation observations are allocated equally among three alternatives in all the cases.

\begin{table}[htb]
\centering
\small
\renewcommand{\arraystretch}{1.5}
\caption{\label{tab:paras7}Parameter settings for different confidence scenarios.}
\begin{tabular}{c|ccc}
\hline
Scenario	&	$\mu^{true}$	&	$(\sigma^{true})^2$	&	$T$	\\\hline
Low-confidence	&	[0.01, 0.02, 0.03]	&	[1,1,1]	&	60	\\
Medium-confidence	&	[0.1, 0.2, 0.3]	&	[1,1,1]	&	60	\\
High-confidence	&	[1, 2, 3]	&	[1,1,1]	&	60\\
\hline
\end{tabular}
\end{table}

\begin{figure}[htbp]
    \centering
        \subfigure[Low-confidence scenario]{
        \includegraphics[width=2in]{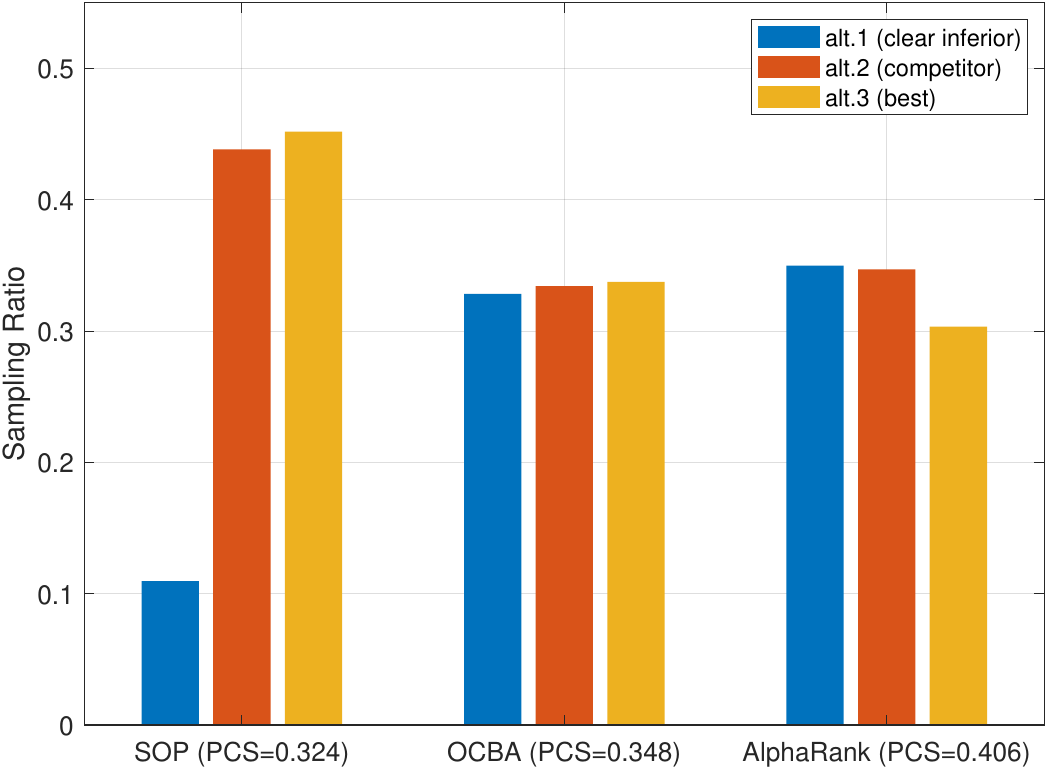}
    }
    \subfigure[Medium-confidence scenario]{
	\includegraphics[width=2in]{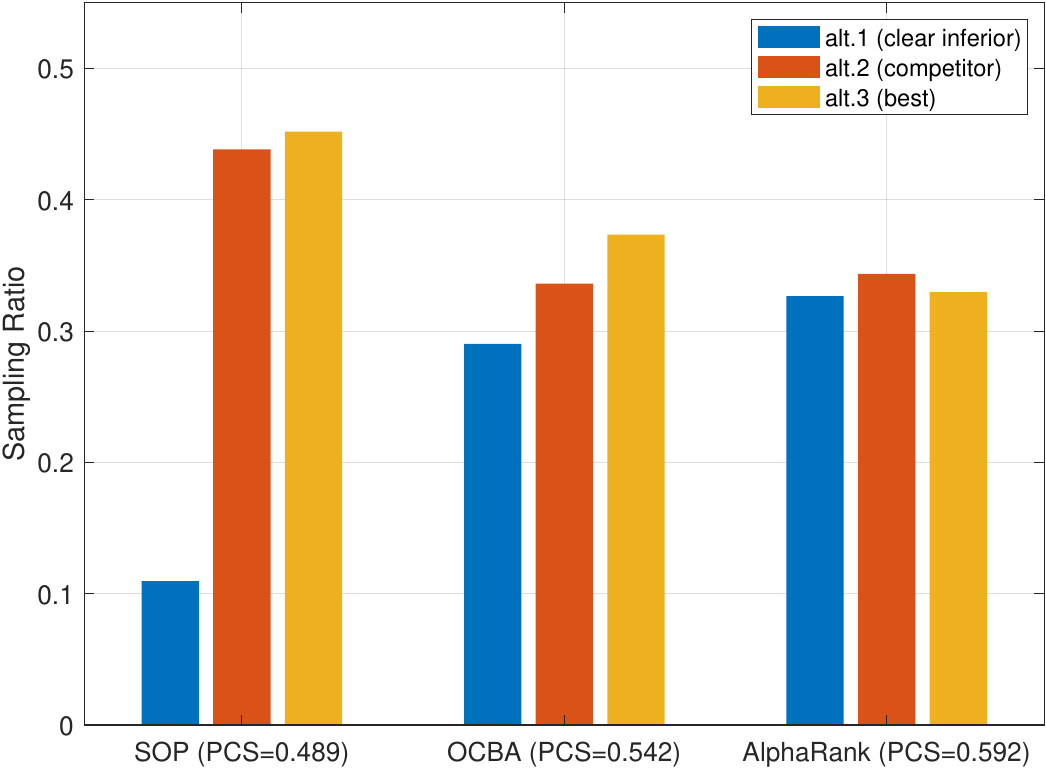}
    }   
    \subfigure[High-confidence scenario]{
	\includegraphics[width=2in]{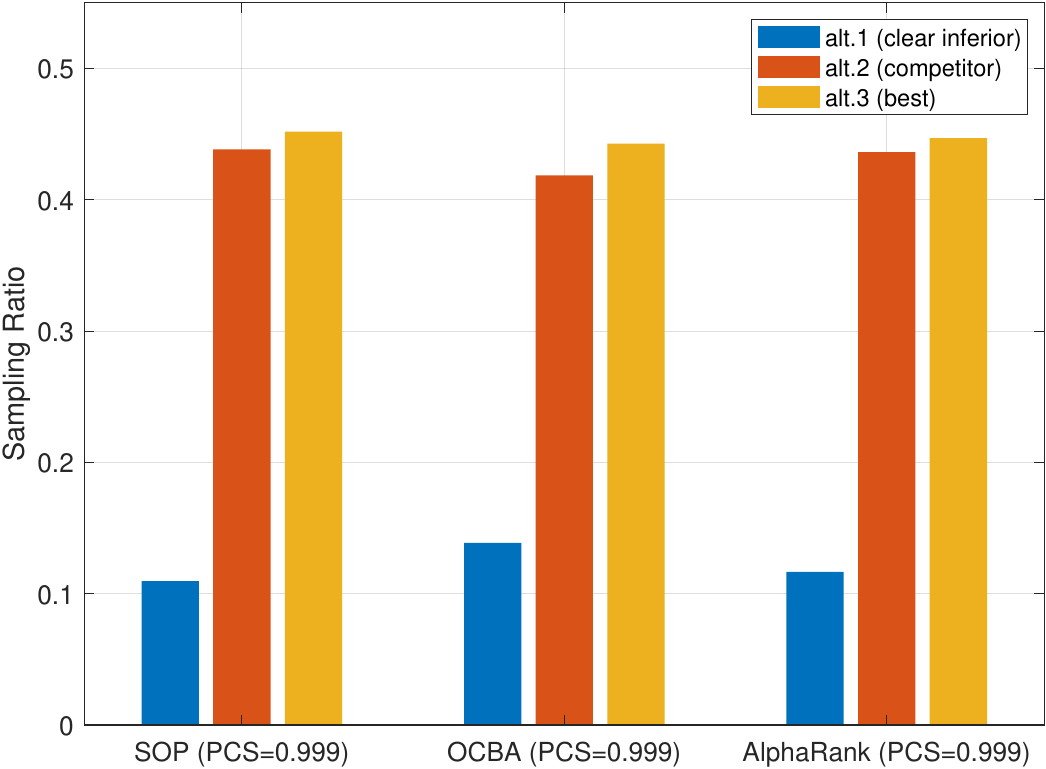}
    }   
    \caption{Sampling ratios of SOP, OCBA, and AlphaRank in different confidence scenarios.}
    \label{ratio3}
\end{figure}

In three confidence scenarios, the sampling behaviors of SOP are almost identical, and both SOP and OCBA favor allocating alternatives with larger means. From Figure \ref{ratio3}(a), in the low-confidence scenario, AlphaRank allocates most simulation observations to the alternative with the smallest mean, which is opposite to the sampling behavior of SOP and OCBA. Figure \ref{ratio3}(b) shows that in the medium-confidence scenario, AlphaRank allocates slightly more simulation observations to the second best (competior), whereas both SOP and OCBA allocate the best alternative most. Interestingly, we can see in Figure \ref{ratio3}(c) that in the high-confidence scenario, the behaviors of AlphaRank, OCBA, and SOP are almost consistent.
\begin{table}[htb]
\centering
\small
\caption{\label{tab:exx2}PCSs of SOP, OCBA, and AlphaRank in different confidence scenarios.}
\begin{tabular}{c|ccc}
\hline
Scenario	&	SOP	&	OCBA	& AlphaRank	\\\hline
Low-confidence	&	0.324	&	0.348	&	0.406	\\
Medium-confidence	&	0.489	&	0.542	&	0.592	\\
High-confidence	&	0.999	&	0.999	&	0.999\\
\hline
\end{tabular}
\end{table}

As indicated in Table \ref{tab:exx2}, AlphaRank consistently achieves the highest PCS in both low and medium-confidence scenarios, while SOP reaches the lowest.   Notably, in high-confidence scenarios, the PCSs of SOP, OCBA, and AlphaRank converge closely to 1, reflecting a diminishing difference in their effectiveness as the confidence increases. In this experiment, we can see that the PCSs and sampling behaviors of  SOP and OCBA get closer to those of AlphaRank from the low-confidence scenario to high-confidence scenario as the induced correlation neglected by SOP and OCBA becomes less significant. AlphaRank can intelligently trade off the information of mean, variance, and induced correlation in different confidence scenarios. 
 \subsection{Sampling Behavior in Different Scales}\label{section84}
With the average number of simulation observations allocated to each alternative $T/N$ being kept constant,  we adjust the scale of the number of alternatives $N$ to observe the change of the sampling ratios.  We set $N$=10, 20, 100 and 1000, and train NN models of scales 10, 20, 100 to directly solve the first three problems. To solve the problem with 1000 alternatives, the DCR framework with the NN model of scale 10 is used. In this example, $\mu_i^{true}=0.1\times i$ and $(\sigma_i^{true})^2=1$, $i=1, 2, \dots,N$. Each alternative is equally allocated 5 initial simulation observations.
\begin{figure}[htbp]
    \centering
    \subfigure[10 alternatives scenario]{
	\includegraphics[width=3in]{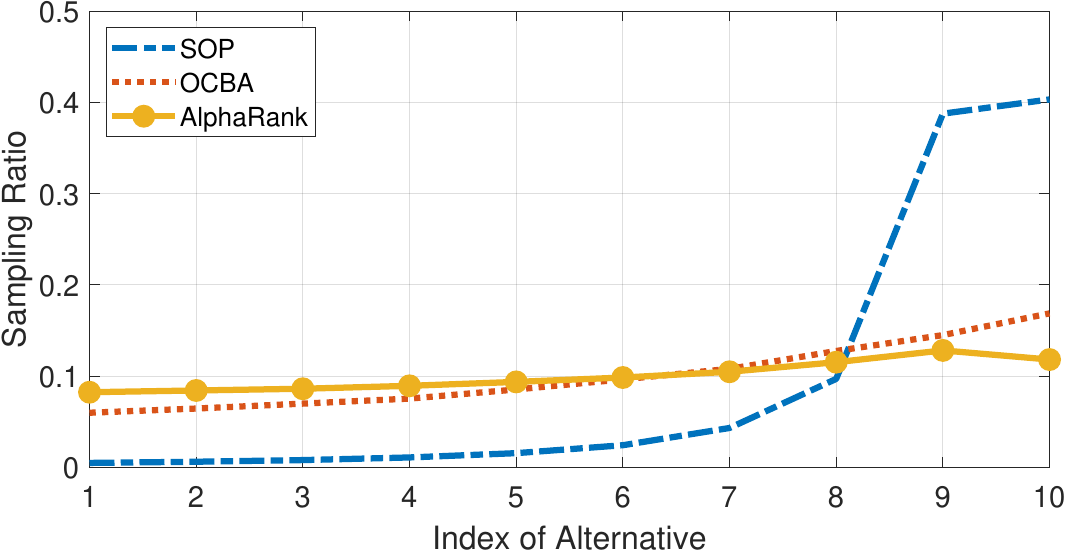}
    }
    \subfigure[20 alternatives scenario]{
	\includegraphics[width=3in]{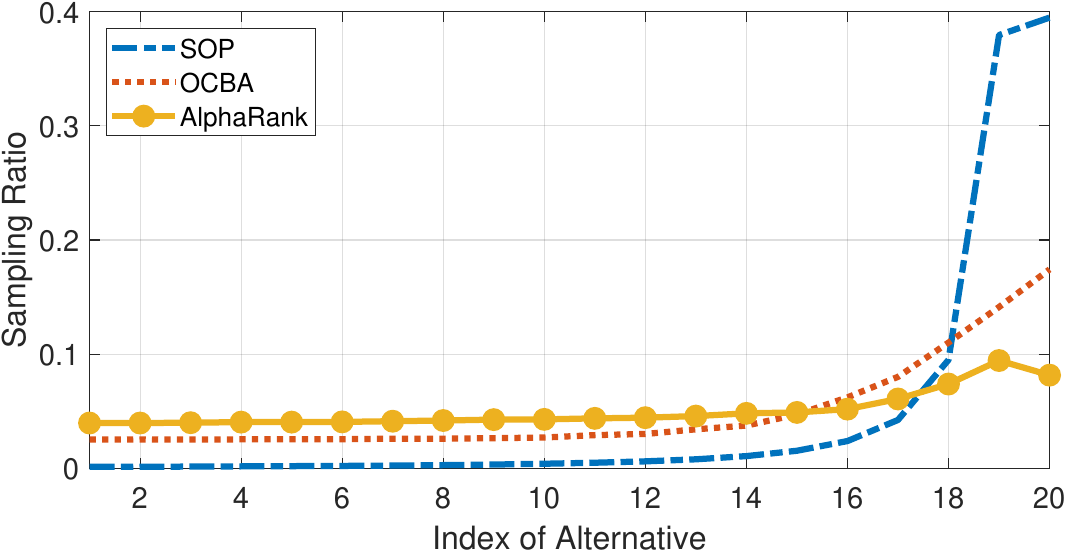}
    }
    \subfigure[100 alternatives scenario]{
	\includegraphics[width=3in]{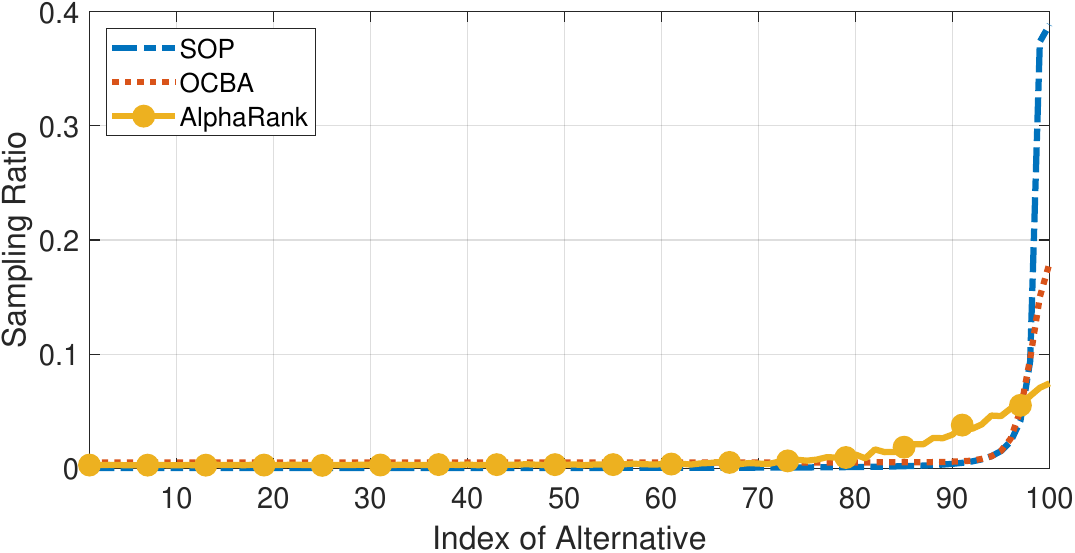}
    }
       \subfigure[1000 alternatives scenario]{
        \includegraphics[width=3in]{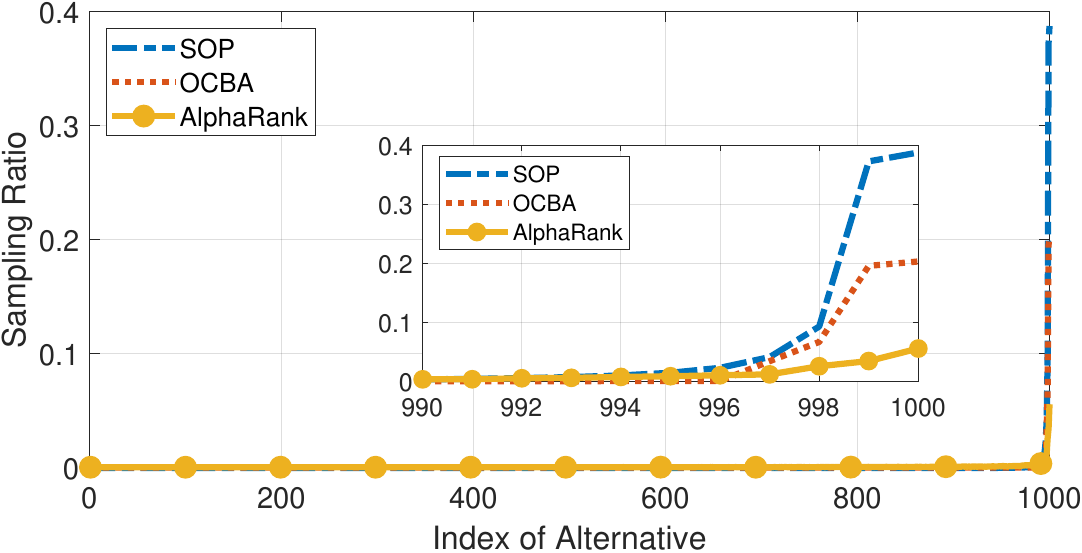}
    }
    \caption{Sampling ratios of SOP, OCBA, and AlphaRank in different scenarios.}
    \label{ratioalt}
\end{figure}

\begin{table}[htb]
\centering
\small
\renewcommand{\arraystretch}{1.5}
\caption{\label{tab:exx3}PCSs of SOP, OCBA, and AlphaRank with different scales.}
\begin{tabular}{c|ccc}
\hline
Scale	&	SOP	&	OCBA	&	AlphaRank	\\\hline
10 alternatives	&	0.118	&	0.448	&	0.515	\\
20 alternatives	&	0.103	&	0.527	&	0.605\\
100 alternatives&	0.592	&	0.683	&	0.808	\\
1000 alternatives	&	0.711	&	0.759	&	0.845	\\
\hline
\end{tabular}
\end{table}

From Figure \ref{ratioalt}, both OCBA and SOP consistently favor allocating alternatives with larger means. In smaller to medium-sized R\&S problems, encompassing 10 and 20 alternatives, AlphaRank demonstrates a distinct strategy by allocating more simulation observations to both the best and second-best alternatives.  Notably, the second-best often receives the highest sampling ratio, reflecting AlphaRank is a nuanced approach in allocating simulation observations. Conversely, in large-scale R\&S problems featuring 100 and 1000 alternatives, AlphaRank's sampling behavior more closely mirrors that of SOP and OCBA. This adaptation involves minimal sampling of clearly inferior alternatives and concentrating most observations on the top 20 and 10 alternatives, respectively, showcasing AlphaRank's flexibility across varying scales. As indicated in Table \ref{tab:exx3}, AlphaRank consistently outperforms both SOP and OCBA across various problem scales, affirming its robustness and effectiveness in diverse R\&S scenarios.
\section{Concluding Remarks}\label{section9}
This study presents AlphaRank, a pioneering AI solution uniquely designed to tackle fixed-budget R\&S problems, to effectively resolve the computational intractability associated with solving SDP, especially when faced with the curse-of-dimensionality in scenarios with large steps and state spaces. Utilizing the DCR framework as proposed, AlphaRank enables efficient resolution of large-scale R\&S problems through the deployment of well-trained, small models, which are particularly suited for parallel computing platforms. Since AlphaRank involves offline pre-training, the trained model can be directly used to efficiently solve the specified problem. Consequently, training diverse NN models tailored to various priors and sharing them via cloud platforms could revolutionize R\&S research, offering a convenient, accessible resource that sets new benchmarks in procedural development. Furthermore, by comparing the sampling behavior of the AI-powered sampling procedure with traditional R\&S procedures, we discover new insight on designing well performed R\&S procedures, i.e., intelligent trade-off among mean, variance, and induced correlation neglected by many existing sampling policies.

Exploring the application of AlphaRank to fixed precision R\&S problems represents a compelling avenue for future research. This could involve adapting the AlphaRank methodology to model fixed-precision R\&S problems as infinite horizon MDPs with stopping times, potentially enhancing PCS efficiency. In addition, the development of RL methods to optimally allocate different constraint indicators  (such as simulation budget in fixed-budget problems and precision in fixed-precision problems) within the DCR framework holds significant potential, offering innovative solutions in both fixed-budget and fixed-precision problem contexts.

\ACKNOWLEDGMENT{This work was supported in part by the National Natural Science Foundation of China (NSFC) under Grants 72250065, 72022001, and 71901003, and by the China Scholarship Council (CSC) under Grant CSC202206010152. A preliminary version of this work has been published in Proceedings of 2023 Winter Simulation Conference~\citep{zhou2023}.}


\bibliographystyle{informs2014} 
\bibliography{name}

\renewcommand\thesection{\Alph{section}}

\setcounter{equation}{0}
\setcounter{algocf}{0}
\setcounter{equation}{0}
\setcounter{section}{0}
\setcounter{lemma}{0}
\setcounter{figure}{0}
\setcounter{assumption}{0}
\begin{APPENDIX}{}
\numberwithin{equation}{section}
\numberwithin{lemma}{section}
\numberwithin{table}{section}
\numberwithin{figure}{section}
\numberwithin{algocf}{section}
\numberwithin{assumption}{section}
\section{Proof of Propositions and Corollarys}
\subsection{Proof of Proposition 1}\label{proof2}
\begin{repeattheorem}[Proposition 1.]
The rollout action value of action $a_{t+1}^{(i)}$ in state $s_t$ is the theoretical PCS of the base policy $\pi$ in state $s_{t+1}^{(i)}$, i.e., $Q(s_t,a_{t+1}^{(i)})=\text{PCS}^{\pi}(s_{t+1}^{(i)})$. Furthermore, define ${\rm Pr}^{improve}\left(s_t\right)$ as the probability of policy improvement based on the rollout policy in $s_t$. Then, when the number of rollouts used for value function estimation is $K$, we have
\begin{gather}
{\rm Pr}^{improve}\left(s_t\right)\geq1-\sum_{i=1}^N\left(1- {\rm Pr}\left(\left.Q_t^{(i)}(s_{t+1}^{(i)})-Q_t^{(a^*)}(s_{t+1}^{(a^*)})\leq 0\right|s_t\right)\right),
\end{gather}
where $a^*=\arg\max_{i=1,\dots,N}Q(s_t,a_{t+1}^{(i)})$. 
\end{repeattheorem}
\proof{Proof of Proposition 1.}
From the definition of the rollout policy value function estimation, it can be obtained that
$$Q_t^{(i)}(s_{t+1}^{(i)})=\frac{1}{K}\sum_{k=1}^K r_{t,k}^{(i)}.$$
To simplify the notation, let $p_{i,t}=Q(s_t,a_{t+1}^{(i)})=\text{PCS}^{\pi}(s_{t+1}^{(i)})$, and then the reward value of $k$-th rollout is $r_{t,k}^{(i)}=1$ or $r_{t,k}^{(i)}=0$, with probabilities of $p_{i,t}$ and $1-p_{i,t}$, respectively, i.e., $r_{t,k}^{(i)}$ follows the Bernoulli distribution with parameter $p_{i,t}$. Repeat $K$ rollouts and define $R_{t,K}^{(i)}=\sum_{k=1}^K r_{t,k}^{(i)}$ as a random variable, and $R_{t,K}^{(i)}\sim(K,p_{i,t})$. We have
$$Q\left(s_t,a_{t+1}^{(i)}\right)=\mathbb{E}\left[\left.\frac{1}{K}R_{t,K}^{(i)}\right|s_t\right]=\frac{1}{K}\mathbb{E}\left[\left.R_{t,K}^{(i)}\right|s_t\right]=p_{i,t}=\text{PCS}^{\pi}(s_{t+1}^{(i)}),$$
i.e., the rollout action value of action $a_{t+1}^{(i)}$ in state $s_t$ is the theoretical PCS of the base policy $\pi$ in state $s_{t+1}^{(i)}$.
Define the optimal action for the largest action value (theoretical PCS) as
$$a^*=\arg\max_{i=1,\dots,N}Q(s_t,a_{t+1}^{(i)}).$$
When the value function estimate of the rollout policy is accurate, we must have 
\begin{gather}\label{pro1eq1}
    Q_t^{(a^*)}(s_{t+1}^{(a^*)})=\frac{1}{K}\sum_{k=1}^K r_{t,k}^{(a^*)}=\frac{1}{K}R_{t,K}^{(a^*)}\geq Q_t^{(i)}(s_{t+1}^{(i)})=\frac{1}{K}\sum_{k=1}^K r_{t,k}^{(i)}=\frac{1}{K}R_{t,K}^{(i)}.
\end{gather}
The action given by the rollout policy to allocate the $(t+1)$-th simulation observation is
$$a_{t+1}^{roll}=\arg\max_{i=1,\dots,N}Q_t^{(i)}(s_{t+1}^{(i)}).$$
When $a_{t+1}^{roll}=a^*$, rollout policy can improve the performance of the base policy, and $\text{Pr}^{improve}\left(s_t\right)$, the probability of policy improvement at step $t$, is
\begin{align}
\text{Pr}^{improve}\left(s_t\right)&=\text{Pr}\left(\left.a_{t+1}^{roll}=a^*\right|s_t\right)\notag
\\&=\text{Pr}\left(Q_t^{(a^*)}(\left.s_{t+1}^{(a^*)})=\max_{i=1,\dots,N}Q_t^{(i)}(s_{t+1}^{(i)})\right|s_t\right)\notag\\&=\text{Pr}\left(\left.R_{t,K}^{(1)}\leq R_{t,K}^{(a^*)},\dots,R_{t,K}^{(N)}\leq R_{t,K}^{(a^*)}\right|s_t\right).\label{pro1eq9}
\end{align}
$R_{t,K}^{(a^*)}$ can be regarded as the number of times that the final selection is correct after $K$ rollouts which could be $0, 1,\dots, K$. 
According to Bonferroni's inequality, the
probability of policy improvement in $s_t$ is
\begin{align}\notag
\text{Pr}^{improve}\left(s_t\right)&=\text{Pr}\left(\left.R_{t,K}^{(1)}-R_{t,K}^{(a^*)}\leq 0,\dots,R_{t,K}^{(N)}-R_{t,K}^{(a^*)}\leq 0\right|s_t\right)\notag
\\&\geq1-\sum_{i=1}^N\left(1- \text{Pr}\left(\left.R_{t,K}^{(i)}-R_{t,K}^{(a^*)}\leq 0\right|s_t\right)\right).\label{pro1eq3}
\end{align}\hfill \Halmos \\

\endproof
\subsection{Proof of Proposition 2}\label{proof3}
\begin{repeattheorem}[Proposition 2.]
   Define the selection given by the rollout policy after allocating $T$ simulation observations as $S_T^{roll}(s_{T})$. If Assumptions \ref{ass6} an \ref{ass7} are satisfied, the rollout policy is consistent, i.e.,
 $$\lim_{T\rightarrow \infty}S_T^{roll}(s_{T})=\arg\max_{i=1,\dots,N}\mu_i^{true},\ a.s.$$
\end{repeattheorem}
\proof{Proof of Proposition 2.}
 The sample-allocation for the $T$-th simulation observation given by the rollout policy is
 $$a_{T}^{roll}=\arg\max_{i=1,\dots,N}Q_{T-1}^{(i)}(s_{T}^{(i)})\approx\arg\max_{i=1,\dots,N}Q(s_{T-1},a_{T}^{(i)}).$$
When the selection policy of rollout policy is selecting the alternative with the largest posterior PCS, i.e., 
\begin{gather}\label{rollselect1}
    S_T^{roll}(s_{T}^{(a_{T}^{roll})})=\arg \max_{i=1,\dots,N}  \text{Pr}\left(\left.\mu_i^{true}>\mu_j^{true},\ j\in \mathcal{I}_i\right|s_T\right),
\end{gather}
where $\mathcal{I}_i$ is the set of remaining $N-1$ indexes except $i$, \cite{2016Dynamic} show that if every alternative will be sampled infinitely as $T\rightarrow \infty$, we have 
$$\text{Pr}\left(\left.\mu_{S^*}^{true}>\mu_j^{true},\ S^*=\arg \max_{i=1,\dots,N}\mu_i^{true}, j\in \mathcal{I}_{S^*}\right|s_T\right)\rightarrow 1,\  a.s.,$$
$$\text{Pr}\left(\left.\mu_j^{true}>\mu_i^{true}, S^*=\arg \max_{i=1,\dots,N}\mu_i^{true}, j\in \mathcal{I}_{S^*}, i\in \mathcal{I}_{j}\right|s_T\right)\rightarrow 0,\  a.s..$$
When the rollout policy approximately selects the alternative with the largest posterior distribution of parameters as the final selection, i.e., 
\begin{gather}\label{rollselect2}
    S_T^{roll}(s_{T}^{(a_{T}^{roll})})=\arg\max_{i=1,\dots,N}\mathbb{E}\left[\left.\mu_{i}^{true}\right|s_T\right],
\end{gather}
if every alternative will be sampled infinitely as
$T\rightarrow \infty$, when $\mathbb{E}[|\mathbb{E}\left[\left.\mu_{i}^{true}\right|s_T\right]|]\leq\mathbb{E}[|\mu_i^{true}|]<\infty$, by Doob’s martingale convergence and consistency theorems, we have
$$\lim_{T\rightarrow \infty}\mathbb{E}\left[\left.\mu_{i}^{true}\right|s_T\right]=\mu_i^{true},\ a.s.,$$
$$\lim_{T\rightarrow \infty}\mathbb{E}\left[\left.\max_{i=1,\dots,N}\mu_{i}^{true}\right|s_T\right]=\max_{i=1,\dots,N}\mu_i^{true},\  a.s..$$
Therefore, regardless of whether selection policy of the rollout policy is (\ref{rollselect1}) or (\ref{rollselect2}), if following the rollout policy $a^{roll}$, every
alternative will be sampled infinitely often almost surely as $T\rightarrow \infty$, the rollout policy is consistent. 

When Assumption \ref{ass7} is satisfied and $T\rightarrow \infty$, there's always a certain probability $\text{Pr}^{improve}\left(s_T\right)>0$ that we have 
\begin{gather}\label{eqpro21}
    \text{PCS}^{roll}(s_T^{(a_T^{roll})})\geq\text{PCS}^{\pi}(s_T^{(a_T^{\pi})}),
\end{gather}
where $\pi$ is the base policy used by rollout policy, $a_{T}^{roll}$ and $a_{T}^{\pi}$ represent the action determined by rollout policy and policy $\pi$ as the allocation policy, $\text{PCS}^{roll}$ and $\text{PCS}^{\pi}$ represent PCS obtained based on the allocation policy $a_{T}^{roll}$ and $a_{T}^{\pi}$, respectively. Let alternative $i$ be sampled $T_i$ times by the allocation policy, with $\sum_{i=1}^NT_i=T$. Assumption \ref{ass5} shows $\pi$ is consistent, i.e., when $T\rightarrow\infty$, $T_i\rightarrow\infty$ , $i=1,2,\dots,N$. \cite{2016Dynamic} prove that when $T_i\rightarrow\infty$, the posterior variances of $\mu_i^{true}$ tend to 0, $i=1,2,\dots,N$, and, consequently,
\begin{gather}\label{eqpro22}
    \text{PCS}^{\pi}(s_T^{(a_T^{\pi})})\rightarrow 1.
\end{gather}
 
Therefore, with (\ref{eqpro21}) and (\ref{eqpro22}), we can draw the conclusion that $\text{PCS}^{roll}(s_T^{(a_T^{roll})})\rightarrow 1$ when $\text{PCS}^{\pi}(s_T^{(a_T^{\pi})})\rightarrow 1$ with probability $\text{Pr}^{improve}\left(s_T\right)$. According to the definition of PCS,
$$\lim_{T\rightarrow\infty}\text{PCS}^{roll}(s_T^{(a_T^{roll})})\rightarrow 1\ \Rightarrow\ \lim_{T\rightarrow\infty}\text{Pr}\left(\left.\mu_{S^{roll}_T}^{true}>\mu_j^{true}, j\in \mathcal{I}_{S^{roll}_T}\right|s_{T}^{(a_T^{roll})}\right)\rightarrow 1.$$

 Suppose the rollout policy is used to guide the allocation of simulation observations, and when $T\rightarrow \infty$, there is an alternative $\hat{i}$, whose sampling number $T_{\hat{i}}<\infty$ and all the other alternatives are sampled infinitely, i.e., $T_i=\infty$, $i\neq \hat{i}$. Based on Assumption \ref{ass6}, when $T_{\hat{i}}<\infty$, $\exists\ \epsilon>0$, s.t., 
\begin{gather}\label{eqpro23}
    \epsilon<\text{Pr}(\mu_{\hat{i}}^{true}>\mu_{S_T^{roll}}^{true}|s_T^{(a_T^{roll})})<1-\epsilon,\ a.s..
\end{gather}
Therefore, we have 
$$\lim_{T\rightarrow\infty}\text{PCS}^{roll}(s_T^{(a_T^{roll})})\nrightarrow 1,\ a.s..$$
It contradicts (\ref{eqpro21}) and (\ref{eqpro22}), so we have $T_{\hat{i}}\rightarrow\infty$, and
$$\lim_{T\rightarrow \infty}S_T^{roll}(s_{T}^{(a_T^{roll})})=\lim_{T\rightarrow \infty}\arg\max_{i=1,\dots,N}\mu_i^{true},$$
i.e.,
$$\lim_{T\rightarrow \infty}S_T^{roll}(s_{T})=\arg\max_{i=1,\dots,N}\mu_i^{true},\ a.s..$$\hfill \Halmos \\
\endproof

\subsection{Proof of Proposition 3}\label{proof5}
\begin{repeattheorem}[Proposition 3.]
 Suppose the maximum number of the steps in the rollout is $H$. Then the computational complexity of the rollout policy is $O(N^2TH)$.
\end{repeattheorem} 
\proof{Proof of Proposition 3}
Each alternative will have $N$ decision nodes at step $t$, $t=1,\dots,T$. Depending on the problem scale and goal setting, the number of steps forward for rollouts in each time step is set differently. There are $N$ decision nodes at each step of the forward rollout. When the number of steps to rollout forward is the remaining simulation budget $T-t$, the computational complexity is calculated as 
$$N(T-1)N+N(T-2)N+\dots+N\cdot2\cdot N+N\cdot1\cdot N=O(N^2T2),$$
and the maximum number of forward steps is $H=T$. When the number of steps in each forward rollout is fixed to a constant $h$, the computational complexity is $$T\cdot N\cdot h \cdot N=O(N^2Th),$$
and the maximum number of forward steps is $H=h$. Therefore, the computational complexity of the rollout algorithm is $O(N^2TH)$.\hfill \Halmos \\
\endproof
\subsection{Proof of Proposition 4}\label{proof6}
\begin{repeattheorem}[Proposition 4.]
    Suppose the number of alternative statistics is $C$, the width of the hidden layer of NN is $W$, the number of hidden layers is $n_h$, the number of alternatives is $N$, and the simulation budget is $T$. Then the computational complexity of NN training is $O((CN+1)W+n_{h}W^2+NW)$. The computational complexity of using the NN as the allocation policy in the R\&S problem is $O(NT)$.
\end{repeattheorem}
\proof{Proof of Proposition 4}
To compute the computational complexity of NN training, we need to analyze the computational cost of passing one observation through each module of the NN of AlphaRank, which is obtained by multiplying the input and output dimensions of each layer of the network and then adding them together, i.e., 
$$\underbrace{(CN+1)W}_{input\ layer}+\underbrace{n_{h}W^2}_{n_{h}\ hidden\ layers}+\underbrace{NW}_{output\ layer}=O((CN+1)W+n_{h}W^2+NW).$$
For $N$, NN has a computational complexity of $O(N)$. Therefore, for a problem with simulation budget $T$, the computational complexity for $N$ and $T$ of using the NN as the allocation policy in the R\&S problem is $O(NT)$.\hfill \Halmos \\
\endproof
\subsection{Proof of Corollary 1}\label{proof7}
\begin{repeattheorem}[Corollary 1.]
When a pre-trained NN model for $M$ alternatives is used to guide the allocation policy to solve the problem with $N$ alternatives under the DCR framework, the computational complexities of rollout policy and NN training in the presence of the number of alternatives are $O(M^2)$ and $O(M)$, respectively. The computational complexity of diectly using NN as the allocation policy is $O(\lceil\log_M^N \rceil)$.
\end{repeattheorem}
\proof{Proof of Corollary 1}
Under the DCR framework, there are a total of $\lceil\log_M^N \rceil$ rounds of screening, which are conducted sequentially. NN is used for each group in each round, and the screening of each group in each round is independent, so it can be parallelized. According to Proposition 4, the computational complexity of each round is $O(M)$. Therefore, the computational complexity for $N$ is $O(\lceil\log_M^N \rceil)$.\hfill \Halmos \\
\endproof
\section{Particle Filtering}\label{app:general}
In the normal conjugacy case under Assumption \ref{ass1}, 
Section \ref{section22} analytically expresses how to update posterior distribution of parameter. Subsequently, we expand this method to encompass general parameter distributions, maintaining the assumption of a normal sampling distribution, thus showcasing the method's adaptability across different statistical contexts. The distribution $F_p(\cdot|\varepsilon_t)$ of the current state $s_t$ need to be updated recursively based on the available statistics $\varepsilon_t$ by predicting and updating steps. The prediction step is to compute the current state's posterior probability density using the prior information given by $F_p(\cdot|\varepsilon_{t-1})$. The update step is to calculate the posterior probability density using the most recent observations to modify the prior probability density. We employ particle filtering (\citealt{2001sequential}), a Monte Carlo method commonly utilized in state estimation of partially observable Markov decision process (POMDP), to update the posterior distribution of parameters. The particle filtering process, vital in state estimation for POMDPs, unfolds in three key steps: initialization, which involves generating an initial set of particles based on prior knowledge;   prediction, where each particle forecasts the next state according to the system model; and update, which adjusts particle weights based on new observations, followed by importance sampling to refine the particle ensemble to more accurately reflect the posterior distribution.
 Any POMDP can be converted into an equivalent MDP by merging information states, which is referred to as an information-state MDP (ISMDP). Therefore, R\&S can also be described by a POMDP and the posterior distribution is updated as follows (\citealt{2018A}):
\begin{gather}
    \hat{F}_p(\cdot|\varepsilon_t)=\frac{1}{K}\sum_{i=1}^K\mathbf{1}_{\theta_{i,j}^{(t)}}(\cdot)\to \hat{F}_p(\cdot|\varepsilon_{t+1})=\frac{1}{K}\sum_{i=1}^K\mathbf{1}_{\theta_{i,j}^{({t+1})}}(\cdot),
    \label{eqpf1}
\end{gather}
where $K$ is the number of particles, $\theta_{i,j}^{(t)}$ is the $j$-th particle for $\theta_i$ at step $t$, $j=1, 2, \dots,K$, and $\mathbf{1}_x(\cdot)$ is the delta-measure with a mass
on $x$. The particles $\theta_{i,j}^{(t+1)}$ are resampled from  $\theta_{i,j}^{(t)}$ with weights
\begin{gather}
    w_{i,j}=\frac{f_s^{(i)}(X_{i,t};\theta_{i,j}^{(t)})}{\sum_{l=1}^K f_s^{(i)}(X_{i,t};\theta_{i,l}^{(t)})},\ j=1, 2, \dots,K,
    \label{eqpf2}
\end{gather}
where $f_s^{(i)}(\cdot)$ is the density of the sampling distribution of the $i$-th alternative, $i=1,2,\dots,N$. 

The flow of particle filtering is as follows: first, generate the sample particle set $\{\theta_{i,j}^{(t)}\}$ by prior $F$ to initialize the particle set; then, for $j=1,2, \dots, K$, calculate and normalize particle weights $w_{i,j}$ by \eqref{eqpf2} and resample the particle set $\{\theta_{i,j}^{(t)}, w_{i,j}\}$ to create a new particle set $\{\theta_{i,j}^{(t+1)}, \frac{1}{K}\}$; finally, the posterior distribution of parameters $\hat{F}_p(\cdot|\varepsilon_{t+1})$ is obtained by \eqref{eqpf1}. The pseudo-code of the particle filtering can be found in Algorithm B.1.
\begin{algorithm}\label{alg:ALGO} 
\caption{Particle Filtering}  
\LinesNumbered  
\label{alg1}
\KwIn{ prior distribution of parameters $F_p(\cdot|\varepsilon_{t})$, number of alternatives $N$, number of particles $K$}   

\For{ i=1\  \textbf{to}\  N}  
{  
Particle set initialization: generate sample particle set $\{\theta_{i,j}^{(t)}\}$ by prior $F_p$.\\

    \For{j=1\ \textbf{to}\ K} 
    {  
      Calculate and normalize particle weights $ w_{i,j}=\frac{f_s^{(i)}(X_{i,t};\theta_{i,j}^{(t)})}{\sum_{l=1}^K f_s^{(i)}(X_{i,t};\theta_{i,l}^{(t)})}$.\\
      Resample the particle set$\{\theta_{i,j}^{(t)}, w_{i,j}\}$ to get the new particle set $\{\theta_{i,j}^{(t+1)}, \frac{1}{K}\}$.\\
    }  
   $\hat{F}_p(\cdot|\varepsilon_{t+1})=\frac{1}{K}\sum_{i=1}^K\mathbf{1}_{\theta_{i,j}^{({t+1})}}(\cdot) $
}
\KwOut{the estimated distribution $\hat{F}_p(\cdot|\varepsilon_{t+1})$ of the state at time $t$} 
\end{algorithm}

 Appendix \ref{app:gamma} shows the experimental results of updating the posterior distribution using particle filtering.
\section{Supplement of Numerical Results}
\subsection{Comparative Experiment of Selection Policies}\label{APPD:SUPP0}
We provide numerical results for comparison of various selection and allocation policies. We implement five allocation policies: EA, KG, AOAP, OCBA and rollout policy whose base policy is EA. The selection policies tested are the optimal selection policy choosing the largest posterior PCS (\ref{eq2122}), and selection policy choosing the largest posterior mean (\ref{eqselect1}), which are denoted as OPTIMAL and MEAN, respectively. The posterior PCS is estimated from $10^4$ simulation sampling, and the posterior mean is calculated directly. In this experiment, $\mu_i^{true}\sim N(\mu_i^{(0)},(\sigma_i^{(0)})^2)$ and the number of alternatives is $N = 3$. We present two examples representing high and low confidence scenarios, respectively. In both of the examples, the total simulation budget is $T=60$, $\mu_i^{(0)}=0$, and the true variance is $(\sigma_i^{true})^2=1$, whereas in Example 1, $\sigma_i^{(0)}=0.5$, and in Example 2, $\sigma_i^{(0)}=0.001$, $i=1,2,3$. The initial 30 simulation observations are allocated equally to each alternative, and the remaining simulation observations are allocated according to different policies. 

Note that for EA, KG, AOAP, and OCBA, the selection policy does not impact the allocation policy, as the value function within the allocation policy is independent of the selection policy. Conversely, for the rollout policy, the selection policy is also required during rollout process for making selections, and such selections, in turn, influence the action value function within the allocation policy.

\begin{table}[htb]
\centering
\renewcommand{\arraystretch}{1.5}
\small
\caption{\label{tab:PCSapp1}PCS of each method in different examples.}
\begin{tabular}{c|c|ccccc}
\hline
\multirow{3}*{Example 1} & Method &EA & KG & AOAP & OCBA & rollout \\\cline{2-7}
& OPTIMAL &0.8586&	0.8503&	0.8657&	0.8661&	0.9064\\\cline{2-7}
& MEAN &0.8583&	0.8502&	0.8651&	0.8658&	0.9048\\\hline
\multirow{3}*{Example 2} & Method &EA & KG & AOAP & OCBA & rollout \\\cline{2-7}
& OPTIMAL & 0.3881&	0.3866&	0.3989&	0.3912&	0.4894\\\cline{2-7}
& MEAN &0.385&	0.3857&	0.3982&	0.3879&	0.4663 \\\hline
\end{tabular}
\end{table}
\begin{table}[htb]
\centering
\renewcommand{\arraystretch}{1.5}
\small
\caption{\label{tab:EOCapp1}EOC of each method in different examples.}
\begin{tabular}{c|c|ccccc}
\hline
\multirow{3}*{Example 1} & Method &EA & KG & AOAP & OCBA & rollout \\\cline{2-7}
& OPTIMAL &0.0268&	0.0294&	0.0238&	0.0243&	0.0163\\\cline{2-7}
& MEAN &0.0269&	0.0292&	0.0239&	0.0244&	0.0168\\\hline
\multirow{3}*{Example 2} & Method &EA & KG & AOAP & OCBA & rollout \\\cline{2-7}
& OPTIMAL & 0.0232&	0.0232&	0.0225&	0.0226&	0.0189\\\cline{2-7}
& MEAN &0.0233&	0.0233&	0.0226&	0.0228&	0.0199 \\\hline
\end{tabular}
\end{table}
\begin{table}[htb]
\centering
\renewcommand{\arraystretch}{1.5}
\small
\caption{\label{tab:timeapp1}CPU time of each method.}
\begin{tabular}{c|ccccc}
\hline
 Method &EA & KG & AOAP & OCBA & rollout \\\hline
 OPTIMAL &0.0011s&	0.0012s&	0.0027s&	0.0019s&	9.082s\\\hline
 MEAN &0.0003s&	0.0003s&	0.0017s&	0.0011s&	0.837s\\\hline
\end{tabular}
\end{table}

Tables \ref{tab:PCSapp1} and \ref{tab:EOCapp1} show the PCS and EOC of each method in the two examples, respectively. The performance of the OPTIMAL and MEAN selection policies are close but the former is slightly better regardless of the allocation policy. The disparities in PCS and EOC across various selection policies become more pronounced in low-confidence scenarios. This aligns with \cite{2016Dynamic}'s findings that selection policies differ more significantly when true mean differences are smaller, especially under limited simulation budgets. Table \ref{tab:timeapp1}  provides the CPU time required for each method to complete a single simulation (60 simulation observations allocated). While OPTIMAL as a selection policy enhances PCS, especially in low-confidence scenarios, it demands significantly more time than MEAN, particularly evident in the context of the rollout policy. As evident from Tables \ref{tab:PCSapp1}, while the adoption of OPTIMAL in the rollout policy makes an approximately 2\% improvement in PCS in the low-confidence scenario, it comes at a cost of more than tenfold in terms of time. This cost-effectiveness trade-off renders the choice of OPTIMAL less favorable for the rollout policy. Considering the inherently high computational demands of the rollout policy, a strategic approach is necessary to balance performance with computational efficiency.   Hence, adopting MEAN as the selection policy emerges as a more practical choice for the rollout policy, providing a reasonable balance between PCS improvement and computational resource utilization.

\subsection{AlphaRank with 100 Alternatives}\label{APPD:SUPP3}
  In the training of AR100, the simulation budget is T=2000. At each time step, the rollout looks 200 steps forward and it is repeated 100 times. Figure \ref{large} shows the PCS curves of AlphaRank as the number of training rounds increases. The experimental results show that with the training time and the number of rounds increasing, AlphaRank achieves superior performance compared to the base policy in the medium-scale R\&S problem after 30 rounds of training. The final PCSs of AlphaRank with different base policies increase by about 5\% relative to those of the corresponding base policies. 
\begin{figure}[htb]
\begin{center}
\includegraphics[width=3in]{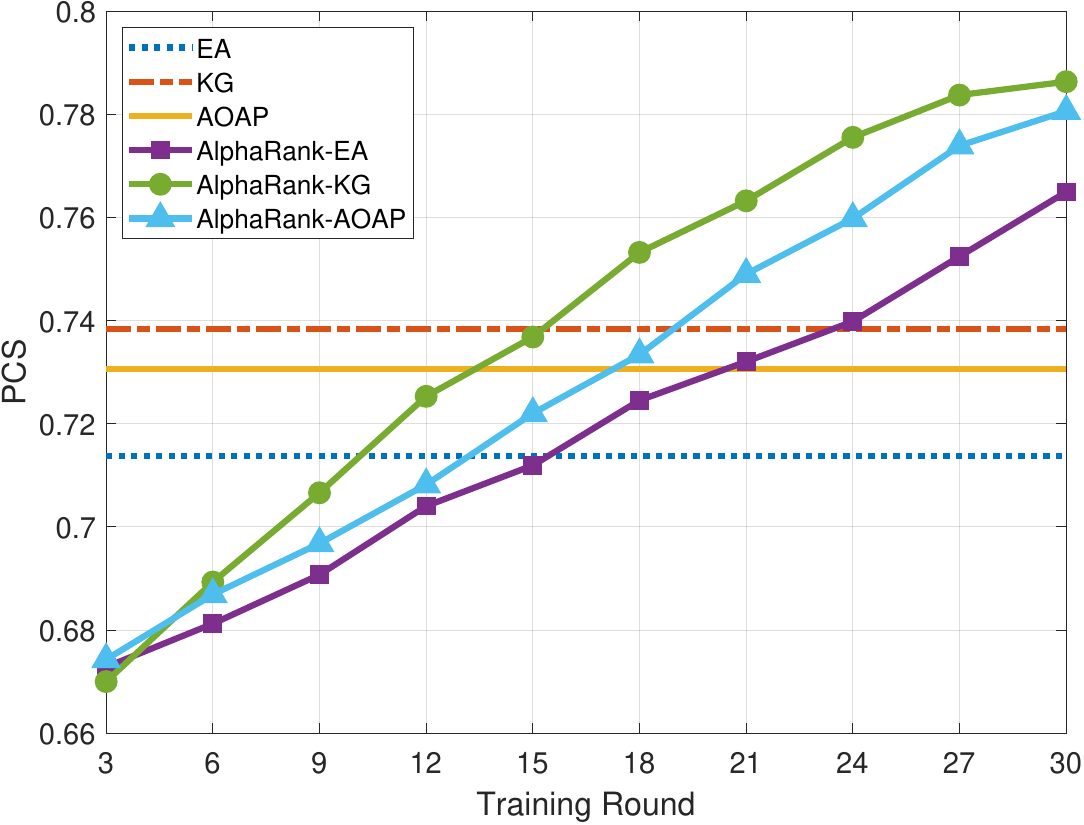}
\caption{Training procedure of value network in parallel.} \label{large}
\end{center}
\end{figure}
\subsection{DCR framework}\label{APPD:SUPP4}
\begin{table}[htb]
\renewcommand{\arraystretch}{1.5}
\centering
\small
\caption{\label{tab:sub}PCS and EOC of different procedures.}
\begin{tabular}{c|cccc}
\hline
Method &$\overline{\text{EOC}_1}$&EOC&$\overline{\text{PCS}_1}$&PCS
\\\hline
EA-DCR& 0.0618& 0.0288&	0.6931& 0.7313\\
EA& -& 0.0932& -& 0.5899 \\
AR-EA-DCR& 0.0531& 0.0199& 0.7826& 0.8244\\
KG-DCR& 0.0580& 0.0254& 0.7140&0.7633 \\
KG& -& 0.0827& -& 0.6104 \\
AR-KG-DCR& 0.0511& 0.0172& 0.7998& 0.8346\\
AOAP-DCR& 0.0593& 0.0260&0.7107&0.7593 \\
AOAP& -& 0.0884& -& 0.6072 \\
AR-AOAP-DCR& 0.0520& 0.0179& 0.7931& 0.8327\\
\hline
\end{tabular}
\end{table}
When the DCR framework is adopted to solve this problem, 10,000 alternatives are divided into 100 groups in the first round, 100 alternatives selected in the first round are divided into 1 group in the second round, and the best alternative is finally selected. Table \ref{tab:sub} presents the average PCS ($\overline{\text{PCS}_1}$) that the best alternative in each group is selected in the first round, the average EOC ($\overline{\text{EOC}_1}$) that the best alternative in each group is selected in the first round, the final PCS and EOC when EA, KG, AOAP and AR are used in the DCR framework, and the PCS and EOC when EA, KG and AOAP are directly used for sample-allocation. The notations of EA and AR whose base policy in the first training round is EA used in the DCR framework are EA-DCR and AR-EA-DCR, respectively, and the remaining notations KG-DCR, AR-KG-DCR, AOAP-DCR, and AR-AOAP-DCR are expressed following the same rule as above. 

From Table \ref{tab:sub}, the following conclusions can be drawn. The PCS and EOC performance of each algorithm using the DCR framework is significantly better than that without using it. More specifically, the PCS increases by about 15\% and the EOC decreases by about 0.06, because the number of required simulation observations is reduced by eliminating poor-performing alternatives in time through screening.
\subsection{Gamma and Normal-binomial Parameter Distribution}\label{app:gamma}
We test the rollout policy based on different sampling policies by comparing it with classic R\&S approaches. The following three sample-allocation policies are used for comparison: EA that sequentially allocates the alternatives from the first to the last in a cyclical manner; EI with uninformative prior (\citealt{Donald1998Efficient}); PTV that allocates simulation observations to each alternative proportional to its sample variance, implemented sequentially by the ``most starving” rule (\citealt{2015Non}). In this experiment, we provide two examples where $\mu_i^{true}\sim\Gamma(\varphi,\gamma)$ and $\mu_i^{true}\sim N(\mu_{nb}, (\sigma_{nb})^2)+B(N, p)$, respectively, and the posterior distributions are updated by particle filtering. The number of alternatives is $N = 10$ and the total simulated budget is $T = 200$. The initial 100 simulation observations are equally allocated to each alternative. The hyperparameters in the two examples are set as follows. 

\textbf{Example 1 (gamma):} $\varphi=2$, $\gamma=1$, i.e., $\mu^{true} \sim \Gamma(2, 1)$, and $(\sigma_i^{true})^2=1$, $ i=1,2,\dots,10$. 

\textbf{Example 2 (normal-binomial):} $\mu_{nb}=0$, $(\sigma_{nb})^2=0.01$, $p_{nb}=0.2$, i.e., $\mu^{true} \sim N(0, 0.01)+B(10, 0.2)$, and $(\sigma_i^{true})^2=1$, $ i=1,2,\dots,10$.

We apply the sequential importance sampling and resampling (SIR)
algorithm (\citealt{2006Optimal}), a well-known particle filtering algorithm, to update the posterior distribution of $\mu_i^{true}$. To ensure the good performance of the particle filtering process in updating the posterior distribution, it is essential to determine the number of sample particles, denoted as $N_P$. We use root mean squared error (RMSE) to measure the fitting accuracy of the particle filtering.  We conduct experiments with $N_P=50, 100, 500$ in the two normal distribution examples mentioned earlier.  Tables \ref{tab:RMSE} and \ref{tab:cpu3} present the RMSE and CPU time between PCS obtained using conjugate prior and SIR methods to update the posterior distribution, which indicates that the number of particles is generally proportional to the accuracy of the posterior distribution estimation. In the experiments of the main body of the paper, we adopt $N_P=50$ when combining SIR with the aforementioned R\&S policies.

\begin{table}[htb]
\centering
\renewcommand{\arraystretch}{1.5}
\small
\caption{\label{tab:RMSE}RMSE of each method.}
\begin{tabular}{c|c|cccccc}
\hline
\multirow{2}*{High} & Method & EA-SIR50 & EA-SIR100 & EA-SIR500 & PTV-SIR50 & PTV-SIR100 & PTV-SIR500 \\\cline{2-8}
& RMSE & 0.012 & 0.009 & 0.005 & 0.010 & 0.008 & 0.005\\\hline
\multirow{2}*{Low} & Method & EA-SIR50 & EA-SIR100 & EA-SIR500 & PTV-SIR50 & PTV-SIR100 & PTV-SIR500 \\\cline{2-8}
& RMSE & 0.014 & 0.012 & 0.009  &0.015 & 0.012 & 0.010 \\\hline
\end{tabular}
\end{table}
\begin{table}[htb]
\centering
\renewcommand{\arraystretch}{1.5}
\small
\caption{\label{tab:cpu3}CPU time of each method.}
\begin{tabular}{c|c|cccccc}
\hline
\multirow{2}*{High} & Method & EA-SIR50 & EA-SIR100 & EA-SIR500 & PTV-SIR50 & PTV-SIR100 & PTV-SIR500 \\\cline{2-8}
& CPU time & 0.0216s & 0.0235s & 0.1003s & 0.0245s & 0.0369s & 0.1851s \\\hline
\multirow{2}*{Low} & Method & EA-SIR50 & EA-SIR100 & EA-SIR500 & PTV-SIR50 & PTV-SIR100 & PTV-SIR500 \\\cline{2-8}
& CPU time & 0.0092s & 0.0214s & 0.1938s & 0.0153s & 0.0299s & 0.2138s \\\hline
\end{tabular}
\end{table}

\begin{figure}[htbp]
    \centering
    \subfigure[Gamma distribution example]{
        \includegraphics[width=3in]{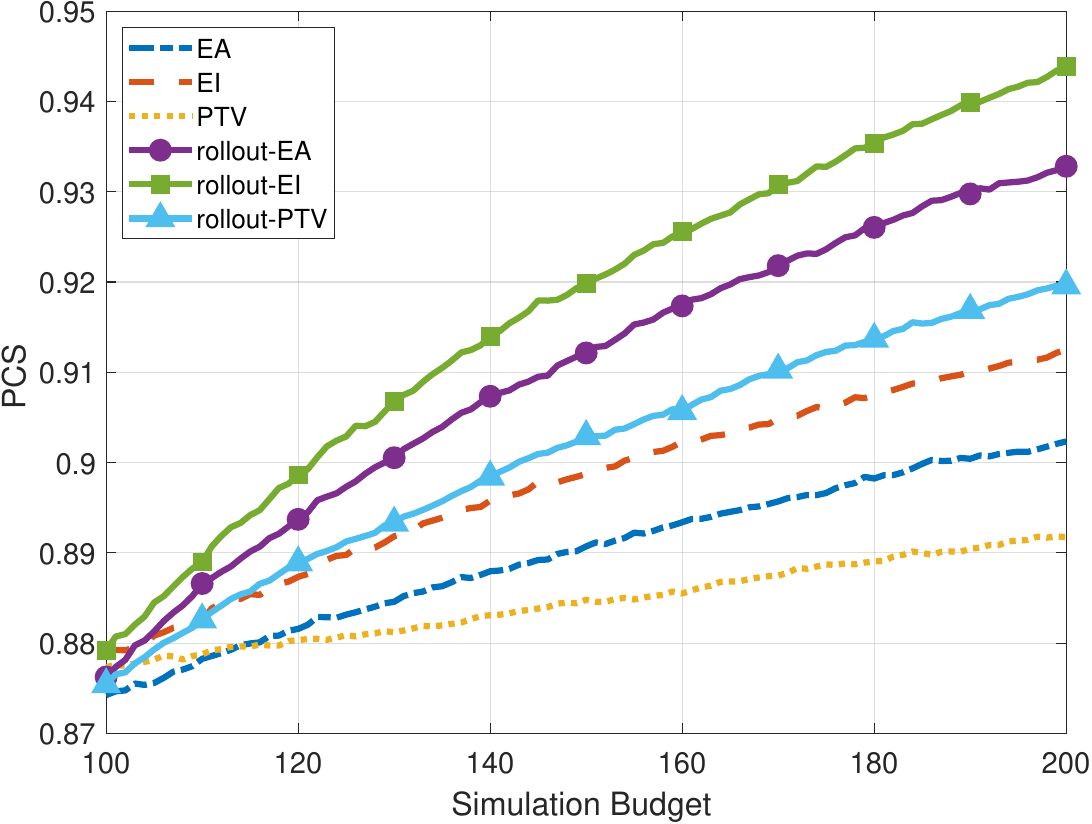}
    }
    \subfigure[Normal-binomial distribution example]{
	\includegraphics[width=3in]{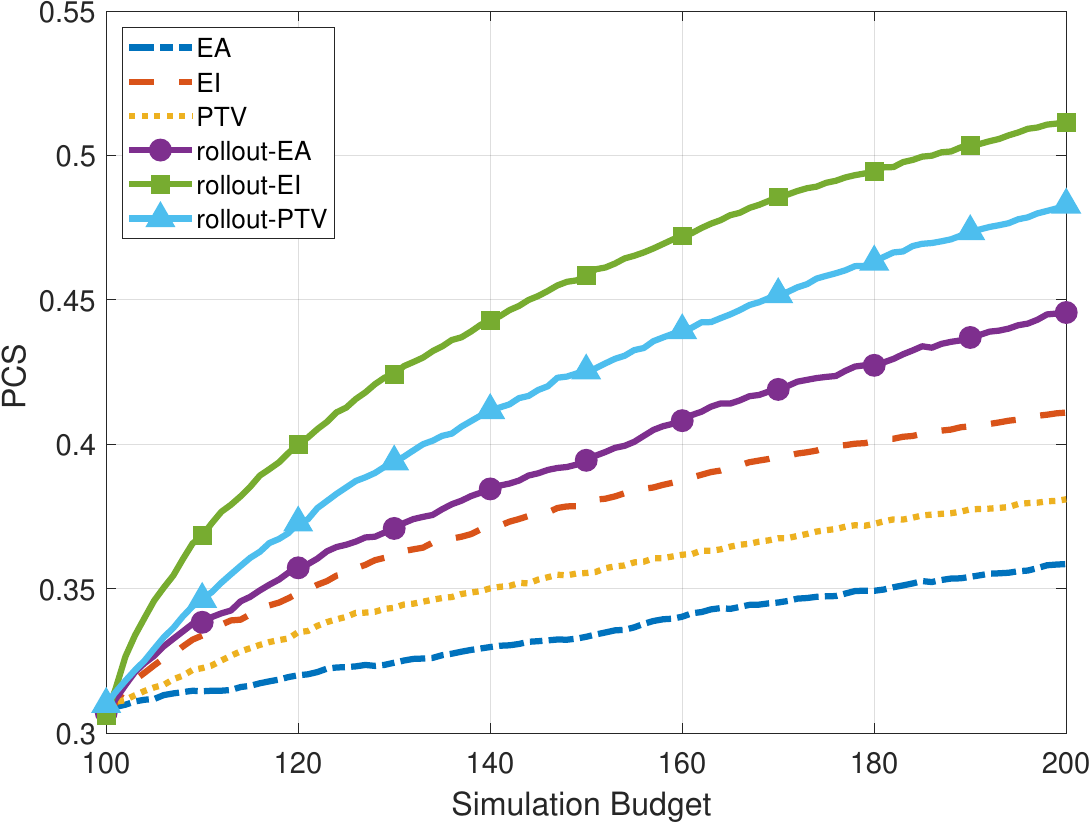}
    }
    \caption{PCSs of EA, EI, PTV, and rollout policies in Experiment 2.}
    \label{roll-general}
\end{figure}
\begin{table}[htb]
\centering
\renewcommand{\arraystretch}{1.5}
\small
\caption{\label{tab:EOC}EOC of each method in different examples.}
\begin{tabular}{c|cccccc}
\hline
Method & EA & EI & PTV & rollout-EA & rollout-EI & rollout-PTV \\\hline
Example 1 & 0.0212 & 0.0172 & 0.0263  &0.0074 & 0.0104 & 0.0134 \\\hline
Example 2 & 0.0855 & 0.0625 & 0.0767  &0.0478 & 0.0212 & 0.0351 \\\hline
\end{tabular}
\end{table}
Figure \ref{roll-general} displays the numerical experiment results evaluating the efficacy of the proposed rollout technique with different base policies and other sampling policies. In Figure \ref{roll-general}(a), the PCSs of EI, EA, and PTV reach a range between 89\% and 91\%, and in  Figure \ref{roll-general}(b), their PCSs range between 35\% and 41\%. Rollout-EA, rollout-EI, and rollout-PTV are significantly superior compared with the corresponding base policies, which improve by about 3\% and 10\% in the two examples, respectively.  
\section{Parameters of Value NN}\label{APPD:PARA}
The following parameters are set for problems with 10 alternatives or less. The number of hidden layers and the number of neurons can be modified according to the complexity of the problem.

\begin{table}[htb]
\renewcommand{\arraystretch}{1.5}
\centering
\small
\caption{\label{tab:net1} Parameters of value NN of normal sampling distribution.}
\begin{tabular}{c|c}
\hline
Layers& Parameters\\\hline
\multirow{2}*{Input Fully Connected Layer} & Number of Layers: 1 \\
~& Width: 4N+1\\\hline
\multirow{2}*{Hidden Fully Connected Layers} & Number of Layers: 3 \\
~& Width: 64\\\hline
\multirow{2}*{Output Fully Connected Layer}& Number of Layers: 1 \\
~& Width: N\\\hline
\end{tabular}
\end{table}

When solving the R\&S problem of general sampling distribution, if the sample mean and sample variance are directly taken as input vectors, the vector length is $2N+1$. Other settings are the same as above.
\end{APPENDIX}	
\end{CJK}
\end{document}